\documentclass[10pt,twocolumn,letterpaper]{article}

\usepackage{cvpr}
\usepackage{times}
\usepackage{epsfig}
\usepackage{graphicx}
\usepackage{amsmath}
\usepackage{amssymb}

\usepackage{epsfig}
\usepackage{amsmath}
\usepackage{amssymb}
\usepackage{makecell}
\usepackage{booktabs}
\usepackage{multirow}
\usepackage{subfigure}

\usepackage{epsfig,float}
\usepackage{picinpar}
\usepackage{graphicx}
\usepackage{epstopdf}
\usepackage{amsmath}
\usepackage{amsthm}
\usepackage{amssymb,bm}
\usepackage[ruled,vlined]{algorithm2e}
\usepackage{float,caption,multirow}
\usepackage{url,caption}
\usepackage{ragged2e}
\usepackage{color}
\usepackage{appendix}

\newcommand{\defeq}{\mathrel{\mathop:}=}

\newcommand{\mc}{\mathcal}
\newcommand{\mb}{\mathbb}

\newtheorem{definition}{Definition}

\usepackage[pagebackref=true,breaklinks=true,letterpaper=true,colorlinks,bookmarks=false]{hyperref}

 \cvprfinalcopy 

\def\cvprPaperID{5187} 

\ifcvprfinal\fi
\begin{document}

\title{LaFIn: Generative Landmark Guided Face Inpainting}

\author{Yang Yang$^1$, Xiaojie Guo$^1$, Jiayi Ma$^{2}$, Lin Ma$^{3}$, and Haibin Ling$^{4}$ \\
	$^1$Tianjin University $^2$Wuhan University  $^3$Tencent AI Lab $^4$Stony Brook University\\
	\tt\small{yangyangcic@tju.edu.cn}, {\{xj.max.guo, jyma2010, haibin.ling\}@gmail.com}, {forestlma@tencent.com}}

\maketitle

\begin{abstract}
   It is challenging to inpaint face images in the wild, due to the large variation of appearance, such as different poses, expressions and occlusions. A good inpainting algorithm should guarantee the realism of output, including the topological structure among eyes, nose and mouth, as well as the attribute consistency on pose, gender, ethnicity, expression, \textit{etc}. This paper studies an effective deep learning based strategy to deal with these issues, which comprises of a facial landmark predicting subnet and an image inpainting subnet. Concretely, given partial observation, the landmark predictor aims to provide the structural information (\textit{e.g.} topological relationship and expression) of incomplete faces, while the inpaintor is to generate plausible appearance (\textit{e.g.} gender and ethnicity) conditioned on the predicted landmarks. Experiments on the CelebA-HQ and CelebA datasets are conducted to reveal the efficacy of our design and, to demonstrate its superiority over state-of-the-art alternatives both qualitatively and quantitatively. In addition, we assume that high-quality completed faces together with their landmarks can be utilized as augmented data to further improve the performance of (any) landmark predictor, which is corroborated by experimental results on the 300W and WFLW datasets. The code is available on \url{https://github.com/YaN9-Y/lafin}
\end{abstract}

\section{Introduction}

\begin{figure}[t]
	\begin{center}
	
	\subfigure{
		\begin{minipage}[t]{0.25\linewidth}
			\includegraphics[width=1in]{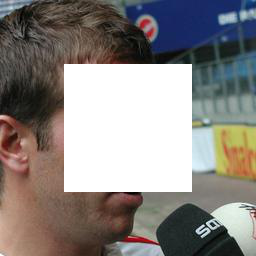}
			\includegraphics[width=1in]{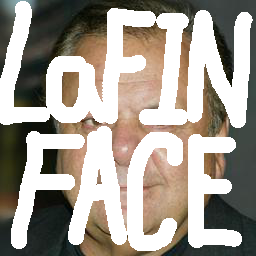}
			\includegraphics[width=1in]{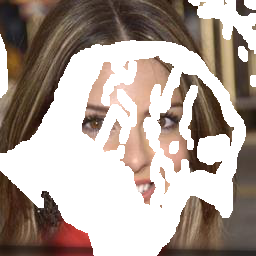} 
		\end{minipage}
	}
	\hspace{.08in}
	\subfigure{
		\begin{minipage}[t]{0.25\linewidth}
			\includegraphics[width=1in]{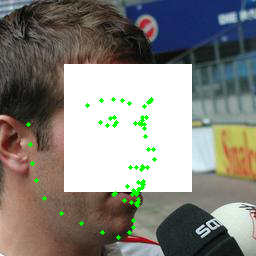}
			\includegraphics[width=1in]{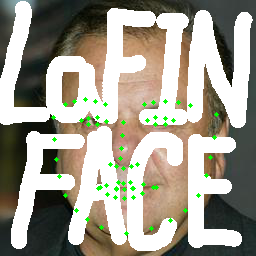}
			\includegraphics[width=1in]{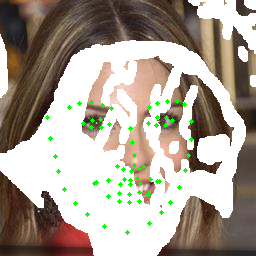}
		\end{minipage}
	}
	\hspace{.08in}
	\subfigure{
		\begin{minipage}[t]{0.25\linewidth}
			\includegraphics[width=1in]{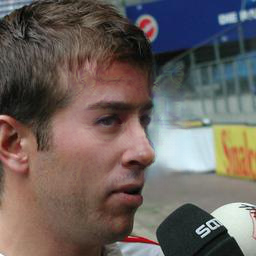}
			\includegraphics[width=1in]{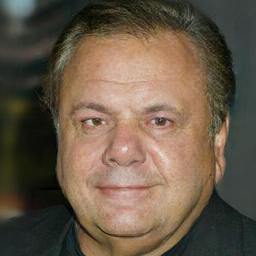}
			\includegraphics[width=1in]{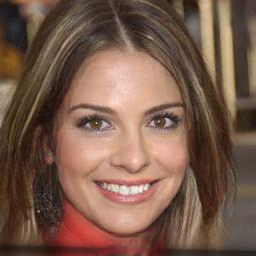}
		\end{minipage}
	}
\end{center}
	\vspace{-10pt}
	\caption{Three face completion results by our method. From left to right: corrupted inputs, plus landmarks predicted from the inputs, and our final results, respectively.}
		\vspace{-7pt}
	\label{fig:open}
\end{figure}

Image inpainting (\textit{a.k.a.} image completion) refers to the process of reconstructing lost or deteriorated regions of images, which can be applied to, as a fundamental component, various tasks such as image restoration and editing  \cite{barnes2009patchmatch,xie2012image}. Undoubtedly, one expects the completed result to be realistic, so that the reconstructed regions can be hardly perceived. Compared with natural scenes like oceans and lawns, manipulating faces, the focus of this work, is more challenging. Because the faces have much stronger topological structure and attribute consistency to preserve. Figure \ref{fig:open} shows three such examples. Very often, given the observed clues, human beings can easily infer what the lost parts possibly, although inexactly, look like. As a consequence, a slight violation on the topological structure and/or the attribute consistency in the reconstructed face highly likely leads to a significant perceptual flaw. The following gives the definition of the problem:
\begin{definition}{Face Inpainting.}
	Given a face image ${I}$ with corrupted regions masked by ${M}$. Let $\overline{{M}}$ designate the complement of ${M}$, and $\circ$ the Hadamard product. The goal is to fill the target part with semantically meaningful and visually continuous information to the observed part.  In other words, the completed result $\hat{{I}}\defeq{M}\circ\hat{{I}}+ \overline{{M}}\circ{I}$ should preserve the topological structure among face components such as eyes, nose and mouth, and the attribute consistency on like pose, gender, ethnicity and expression.
\end{definition}

\subsection{Previous Arts}

Various image inpainting methods have been developed over the last decades. In what follows, we briefly review classic and contemporary works closely related to ours. 

\textit{Traditional Inpainting Methods.} In this category, diffusion-based and patch-based approaches are two representative branches. Diffusion-based approaches  \cite{bertalmio2000image,esedoglu2002digital,yamauchi2003image} iteratively propagate low-level features around the occluded areas. However, these methods are limited to reconstructing structureless and small-size regions. While patch-based methods \cite{barnes2009patchmatch,efros1999texture,huang2014image} attempt to copy similar blocks from either the same image or a set of images to the target regions. On the one hand, their computational cost of calculating the similarity between blocks is expensive, even though some works like \cite{barnes2009patchmatch} have been proposed towards accelerating the procedure. On the other hand, as a common limitation, they all hypothesize that the missing part can be found elsewhere, which does not always hold in practice.

\textit{Deep Learning-based Methods.}
Recently, deep learning based methods have become the mainstream for image inpainting. The context encoder \cite{pathak2016context}, as a pioneer deep-learning method for image completion, introduces an encoder-decoder network trained with an adversarial loss \cite{goodfellow2014generative}. After that, plenty of follow-ups have been proposed to improve the performance from various aspects. For instance, the scheme by \cite{iizuka2017globally} employs both the global and local discriminators to accomplish the task. Another attempt proposed in \cite{yu2018generative} designs a coarse-to-fine network structure and applies a self-attention layer to connect related features at distant spatial locations. Besides,  Yu \textit{et al.} and Liu \textit{et al.} \cite{yu2018free,liu2018image} upgrade the convolutional layers for making networks adaptive to the masked input. However, most of the above-mentioned methods can barely keep the structure of the original image, and the inpainted result frequently tends to be blurry, especially on large occluded areas. For the sake of maintaining the structure of corrupted images, a number of methods, such as \cite{nazeri2019edgeconnect,xiong2019foreground}, try to first predict the edge information for corrupted images, then apply it as a condition to guide the inpainting. These methods work well on small corrupted regions though, when the corruption becomes larger, the performance significantly degrades as it is not easy to predict reasonable edges inside the masked regions, leading to unsatisfactory results. 

\begin{figure}[t]
	\begin{center}
		\includegraphics[width=0.2\linewidth]{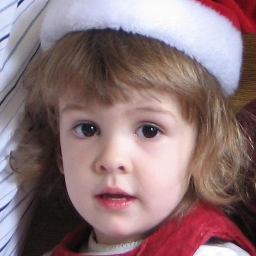}\hspace{-3pt}
		\includegraphics[width=0.2\linewidth]{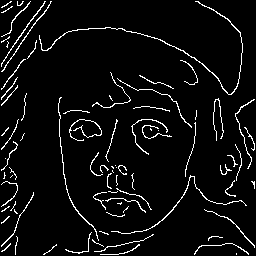}\hspace{-3pt}
		\includegraphics[width=0.2\linewidth]{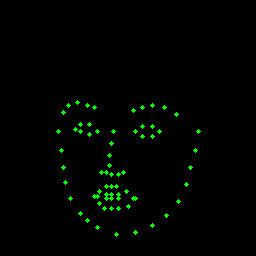}\hspace{-3pt}
		\includegraphics[width=0.2\linewidth]{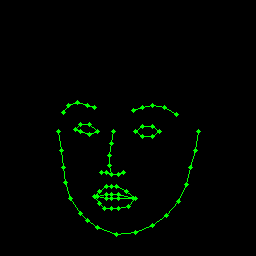}\hspace{-3pt}
		\includegraphics[width=0.2\linewidth]{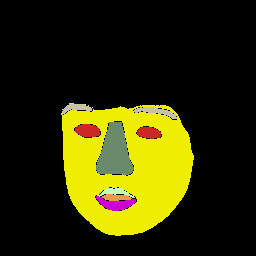}\\
	\end{center}
	\vspace{-7pt}
	\caption{An illustration of different facial features. From left to right: the input, Canny edges, landmarks, edges by connecting the landmarks, and parsing regions.}
		\vspace{-7pt}
	\label{fig:seman}
\end{figure}

\textit{Deep Face Inpainting Methods.}
Specific to face completion, the authors of \cite{GFC} construct a loss taking care of the gap in semantic segmentation (face parsing) between generated face images and ground truth, expecting to better preserve the structure. But this work often suffers from the color inconsistency, and lacks of ability in handling faces with large poses. Besides, \cite{jo2019sc,yu2018free} directly ask users to manually label edges for generating corresponding results. Although providing a flexible way to editing faces, sometimes it is difficult/inconvenient for users to input precise edge information. To relive the requirement from users,  Nazeri \emph{et al.} applied a network to predict the edges \cite{nazeri2019edgeconnect}, which however suffers from inaccurate/unreasonable prediction on large holes. Moreover, we argue that, for face completion, both face parsing and edge information are relatively redundant, which may even degenerate the performance when feeding slightly inaccurate information into the inpainting module. Facial landmarks are better to act as the indicator, which are neat, sufficient, and robust to reflect the structure of face, please see Fig. \ref{fig:seman} for an example. Many works have successfully applied landmarks to the task of face generation, such as \cite{zhang2019faceswapnet}, \cite{headinpaint} and \cite{zakharov2019few}. It is worth noting that, different from the generation task \cite{zhang2019faceswapnet} and \cite{zakharov2019few}, in our problem, the landmarks need to be obtained from the corrupted images. 

\subsection{Challenges and Considerations}

As stated previously, completing face images in the wild is challenging. A qualified face inpainting algorithm should carefully take into account the following two concerns to guarantee the realism of output:
\begin{itemize}
	\item  Faces are of strong structure. The topological relationship among facial features including eyebrows, eyes, nose and mouth is always well-organized. The completed faces must satisfy this topology structure primarily;
	\item The attributes of face, such as pose, gender, ethnicity, and expression, should be consistent across the inpainted regions and the observed part. 
\end{itemize}
Otherwise, a slight violation on these two factors will result in a significant perceptual flaw. 

\begin{figure*}[t]
	\begin{center}
		\includegraphics[width=1\linewidth]{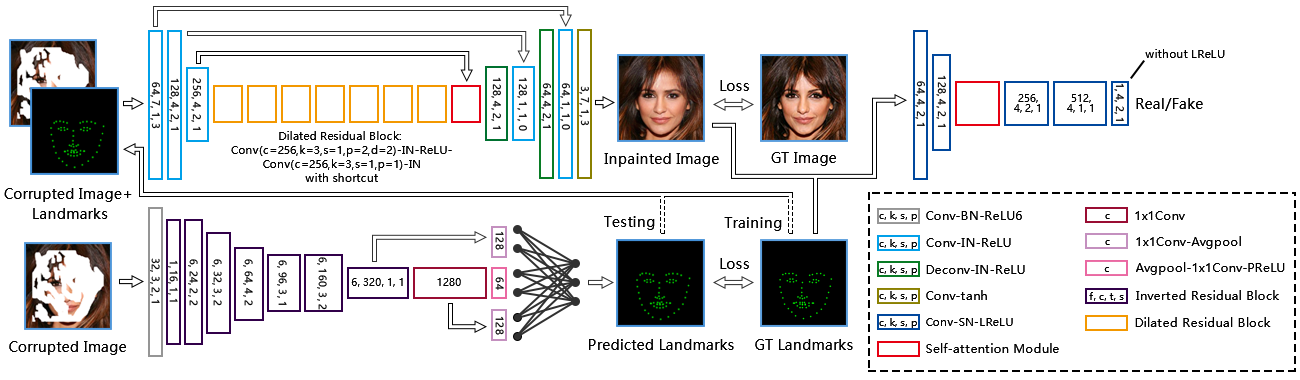}
	\end{center}
	\vspace{-7pt}
	\caption{The architecture of the proposed model. For a corrputed image, the landmarks are first estimated by the landmark prediction module. Then the inpaint module applies the landmarks as prior to inpaint the image. The notations $c$, $k$, $s$ and $p$ stand for the channel number, kernel size, stride, and padding, respectively. In addition, for each inverted residual block, it contains a sequence of identical layers repeating $t$ times,  and the expansion factor is $f$.}
		\vspace{-7pt}
	\label{fig:arch}
\end{figure*}

\textit{Why adopt landmarks?} This work employs facial landmarks as structural guidance, because of  their compactness, sufficiency, and robustness. One may ask whether the edge or parsing information provide more powerful guidance than the landmarks?  If the information is precise, the answer is yes. But, taking the strategy using edges \cite{nazeri2019edgeconnect} as an example, it is not easy to generate reasonable edges in challenging situations like large-area corrupted faces with large-poses. Under the circumstances, the redundant and inaccurate information would instead hurt the performance. Alternatively, a set of landmarks (pre-defined fiducial points) always exists, no matter what situation the face is in. Further, the landmarks can be viewed as the discrete/ordered points sampled on the \textbf{key} edges/regions of face, which are sufficient to conversely reform the \textbf{key} edges/facial regions (face parsing) with redundant information removed. Compared with the edge \cite{nazeri2019edgeconnect} and parsing \cite{GFC} information, for one thing, the landmarks are much neater and more robust, please see Fig. \ref{fig:seman} for illustration. For another thing, once the landmarks for a face are obtained, the topology structure, pose and expression can be subsequently determined. Moreover, the landmarks are more convenient to control from the editing perspective. These properties support that using landmarks is a better choice for face completion.

\textit{How to guarantee the attribute consistency?} Except for the pose and expression attributes determined by the landmarks, there are several other attributes, such as gender, ethnicity, and makeup style, need to be concerned. Notice that the consistency is to bridge the observed and the inpainted regions. This is to say, for these finer-grained attributes, the inpainting algorithm should take the observed (real) information as reference for the reconstruction. Harnessing distant spatial context (large receptive field) and connecting temporal feature maps (long-short term) can effectively fulfill the requirement.  

\subsection{Contributions}
This paper presents a deep network, namely Generative \textbf{La}ndmark Guided \textbf{F}ace \textbf{In}paintor ({LaFIn} for short), which comprises of a facial landmark predicting subnet and an image inpainting subnet, for solving the face inpainting problem. The main contributions can be summarized in the following aspects.

\begin{itemize}
	\item As analyzed, facial landmarks are neat, sufficient, and robust to act as the indicator for face inpainting. We construct a module for predicting landmarks on incomplete faces, which reflect the topological structure, pose and expression of the target face.    
	\item To complete faces, we design an inpainting subnet that employs the predicted landmarks as guidance. For the attribute consistency, the subnet harnesses distant spatial context and connects temporal feature maps.
	\item Extensive experiments are conducted to reveal the efficacy of our design and, show its advances over state-of-the-art alternatives both qualitatively and quantitatively.
\end{itemize}

In addition, we can further use the completion results to help boosting the performance of data-driven landmark detectors. Since, in real situations, the training data are often insufficient, and manually labeling landmarks  is  time-consuming, a simple yet reliable data augmentation manner is definitely desired. Our another contribution is as follows: 
\begin{itemize}
	\item The completion can generate various plausible new faces \textbf{conditioned} on the landmarks. Thus, the generated face and the corresponding (ground-truth) landmarks can be employed as the augmented data to relieve the pressure from manual annotation. The effectiveness of this manner is confirmed by experimental results on the WFLW and 300W datasets.     
\end{itemize}

\section{Methodology}

A desired face inpaintor should generate natural-looking results with logical structures and attributes. We build a deep network, denoted as LaFIn, to achieve the goal. As schematically illustrated in Fig. \ref{fig:arch}, the network is composed of a subnet for predicting landmarks, and another one for generating new pixels conditioned on the predicted landmarks. In the next subsections, we shall detail the network.

\subsection{Landmark Prediction Module}

The landmark prediction module $\mc{G}_L$ aims to retrieve a set of ($n=68$ in this work) landmarks from a corrupted face image ${I}^{M}\defeq{I}\circ{M}$, \textit{i.e.} $\hat{{L}}\in\mb{R}^{2\times n}\defeq \mc{G}_{L} \left({I}^{M}; \theta_{L}\right)$, with $\theta_{L}$ the trainable parameters. 
Technically, $\mc{G}_{L}$ can be accomplished by any landmark detector like \cite{RDR,PCD-CNN,LAB}. Please notice that, for the target inpainting task, what we expect from the landmarks is more about the underlying topology structure and some attributes (pose and expression) than the precise location of each individual landmark. The following may explain the reason: considering the landmarks on face contour for an example, with the corresponding region fixed, shifting them along the contour will not affect the final result much. Consequently, we build a simple yet sufficiently effective $\mc{G}_L$. Our $\mc{G}_{L}$ is built upon the MobileNet-V2 model proposed in \cite{sandler2018mobilenetv2}, which focuses on feature extraction. The final landmark prediction is achieved by fully connecting the fused feature maps at different rear stages, as illustrated in Fig. \ref{fig:arch}. The training loss for $\mc{G}_L$  is simply as follows:
\begin{equation}
\mc{L}_{lmk}\defeq\|\hat{L} - {L}_{gt}\|^2_{2},
\end{equation}
where ${ L}_{gt}$ denotes the ground-truth landmarks. In addition, $\|\cdot\|_2$ stands for the $\ell_2$ norm. 

\subsection{Image Inpainting Module}
The inpainting network $\mc{G}_{P}$ desires to complete faces by taking corrupted images $I^M$ and their (predicted or ground-truth) landmarks $L$ ($\hat{L}$ or $L_{gt}$) as input, \textit{i.e.} $\hat{{I}}\defeq \mc{G}_{P} \left({I}^{M}, L; \theta_{P}\right)$, with $\theta_{P}$ the network parameters. This subnet comprises of a generator and a discriminator.

\textit{Generator.} Overall, the generator is based on the U-Net structure. More specifically, the network consists of three gradually down-sampled encoding blocks, followed by $7$ residual blocks with dilated convolutions and a long-short term attention block. Then, the decoder processes the feature maps gradually up-sampled to the same size as input. The long-short attention layer  \cite{zheng2019pluralistic} is harnessed to connect temporal feature maps, and the stacked dilated blocks are to enlarge the receptive filed so that features in a wider range can be taken into account. Besides, shortcuts are added between corresponding encoder and decoder layers. Moreover, the $1\times1$ convolution operation is executed before each decoding layer as the channel attention to adjust weights of features from the shortcut and last layer. In such a way, the network can better make use of distant features both spatially and temporally. The structure of the generator can be found in Fig. \ref{fig:arch}, and more in Appendix.

\textit{Discriminator.} Based on the concept of two-player game, the generator tries to produce completed faces conditioned on the landmarks to fool the discriminator, while the discriminator aims to determine whether the generated result satisfies the data distribution. The convergence is reached when the generated results are not distinguishable from the real ones.  In this work, our discriminator is built upon the  $70\times70$ Patch-GAN architecture \cite{isola2017image}. To stabilize the training process, we introduce the spectral normalization (SN) \cite{miyato2018spectral} into the blocks of the discriminator. Besides, an attention layer is inserted to adaptively treat the features. It is worth to notice that the works like \cite{iizuka2017globally} employ two discriminators, \textit{i.e.} a global discriminator focuses on the entire image to assess if it is coherent as a whole, and a local one looks only at the completed region to ensure the local consistency. Differently, our discriminator adopts only one judger to accomplish the job, which takes an image and its landmarks as input, \textit{i.e.} $\mc{D}(I, L;\theta_D)$ with $\theta_D$ the parameters. The reasons are: 1) the generated results are conditioned on the landmarks, already ensuring the global structure;  and 2) the attention layer concentrates more on the attribute consistency. The configuration of our discriminator can be found in Fig. \ref{fig:arch}, and more details in Appendix.

\textit{Loss.} We use a combination of a per-pixel loss, a perceptual loss , a style loss, a total variation loss and an adversarial loss,  for training the inpaintor. \\
\noindent\textbf{(I)} {The per-pixel loss} is defined as follows:
\begin{equation}
{\mathcal{L}}_{pixel} \defeq \frac{1}{N_m} \| \hat{I} - { I}\|_1, 
\end{equation}
where  $\|\cdot\|_1 $ stands for the $\ell_1$ norm. Notice that we use the mask size $N_m$ as the denominator to adjust the penalty. It means that if a face is interfered by a small occlusion, the inpainted result should be very close to the ground-truth, while if the corruption is large, the restriction can be relaxed as long as the structure and consistency are rational.\\
\noindent\textbf{(II)} {The perceptual loss} measures the difference of feature maps extracted from a pre-trained network, which is calculated in the following manner:
\begin{equation}
{\mathcal{L}}_{perc} \defeq \sum_p \frac{\|\phi_p(\hat{I}) -  \phi_p\left({I}\right) \|_1}{N_p\times H_p\times\ W_p},
\end{equation}
where $\phi_{p}(\cdot)$ denotes the $N_p$ feature maps with size $H_p\times W_p$ of the $p$-th layer from the pre-trained network. $relu1\_1$, $relu2\_1$, $relu3\_1$, $relu 4\_1$ and $relu5\_1$ of the VGG-19 pre-trained on the ImageNet \cite{russakovsky2015imagenet} are utilized to calculated the perceptual loss, as well as the style loss described below.\\
\noindent\textbf{(III)} {The style loss} computes the style distance between two images as follows:
\begin{equation}
\mathcal{L}_{style} \defeq \sum_p\frac{1}{N_p \times N_p}\|\frac{G_p(\hat{I}\circ {M}) - G_p({I}\circ{M})}{N_p\times H_p\times W_p}\|_1,
\end{equation}
where $G_p(x) = \phi_p(x)^T\phi_p(x)$ stands for the Gram Matrix corresponding to $\phi_p(x)$.\\
\noindent\textbf{(IV)} {The total variation loss} is utilized to suppress the checkerboard artifact, which is defined as:
\begin{equation}
\mathcal{L}_{tv} \defeq \frac{1}{{N_{I}}}\|\nabla \hat{I}\|_1,
\end{equation}
where $N_I$ is the pixel number of $I$, and $\nabla$ is the first order derivative, containing $\nabla_h$ (horizontal) and $\nabla_v$ (vertical).\\
\noindent\textbf{(V)} {The adversarial loss} adopts the LSGAN proposed in \cite{mao2017least}, due to its stability during the training process and the advance in visual quality, which is as follows:
\begin{equation}
\begin{aligned}
{\mathcal{L}}_{adv_G} &\defeq \mathbb{E}[(\mc{D}(\mc{G}_P(I^M,L),L_{gt})-1)^2],\\
{\mathcal{L}}_{adv_D}&\defeq \mathbb{E}[\mc{D}(\hat{I}, L_{gt})^2] + \mathbb{E}[(\mc{D}(I, L_{gt})-1)^2].
\end{aligned}
\end{equation}

\noindent{The total loss} with respect to the generator yields:
\begin{equation}
\begin{aligned}
{\mathcal{L}}_{inp} \defeq  \mathcal{L}_{pixel}& + \lambda_{perc}\mathcal{L}_{perc}+\lambda_{sty}\mathcal{L}_{style} \\
 &+ \lambda_{tv}\mathcal{L}_{tv} + \lambda_{adv}\mathcal{L}_{adv_G}.
\end{aligned}
\end{equation}
We use $\lambda_{perc}=0.1$, $\lambda_{style}=250$, $\lambda_{tv}=0.1$ and $\lambda_{adv}=0.01$ in our experiments. The whole training procedure alternatively minimizes $\mathcal{L}_{inp}$ for the generator $\mc{G}_{P}$ and  ${\mathcal{L}}_{adv_D}$ for the discriminator $\mc{D}$ until converged.\\

\begin{figure*}[t]
	\begin{center}
		\subfigure[Ground truth (GT)]{
			\begin{minipage}[t]{0.15\linewidth}
				\includegraphics[width=1.13in]{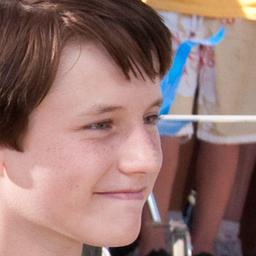}
				\includegraphics[width=1.13in]{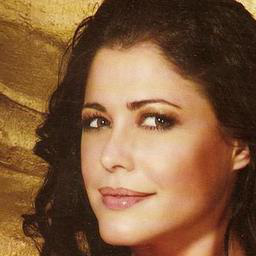}
			\end{minipage}
		}
		\subfigure[Masked GT]{
			\begin{minipage}[t]{0.15\linewidth}
				\includegraphics[width=1.13in]{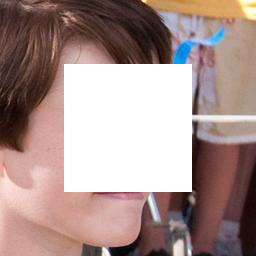}
				\includegraphics[width=1.13in]{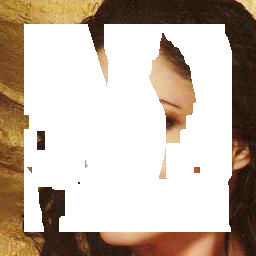}
			\end{minipage}
		}
		\subfigure[CA]{
			\begin{minipage}[t]{0.15\linewidth}
				\includegraphics[width=1.13in]{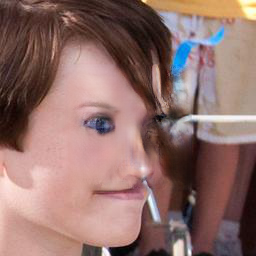}
				\includegraphics[width=1.13in]{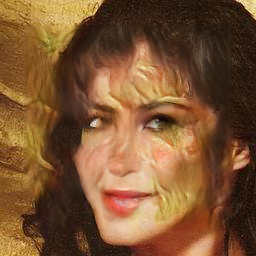}
			\end{minipage}
		}
		\subfigure[EC]{
			\begin{minipage}[t]{0.15\linewidth}
				\includegraphics[width=1.13in]{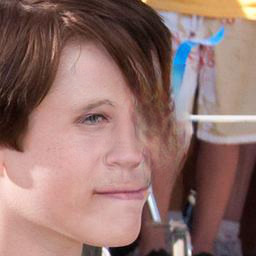}
				\includegraphics[width=1.13in]{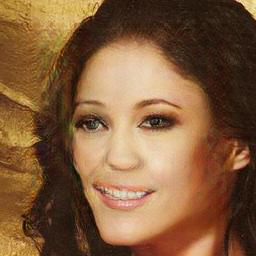}
			\end{minipage}
		}
		\subfigure[PIC]{
			\begin{minipage}[t]{0.15\linewidth}
				\includegraphics[width=1.13in]{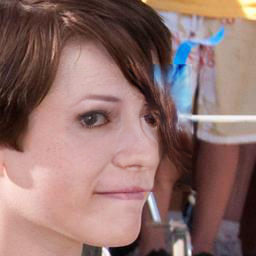}
				\includegraphics[width=1.13in]{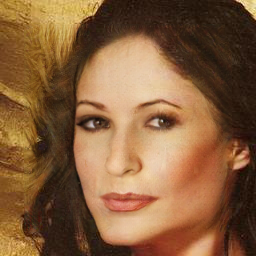}
			\end{minipage}
		}
		\subfigure[Ours]{
			\begin{minipage}[t]{0.15\linewidth}
				\includegraphics[width=1.13in]{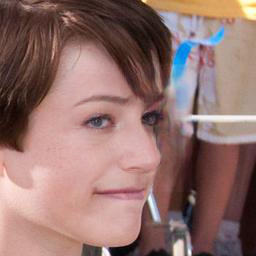}
				\includegraphics[width=1.13in]{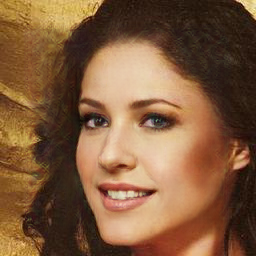}
			\end{minipage}
		}
	\end{center}
\vspace{-15pt}
	\caption{Qualitative comparison with other state-of-the-art techniques on the CelebA-HQ dataset. (a) shows the ground-truth images. (b) depicts the masked versions of (a). (c)-(f) are the results obtained by CA, EC, PIC, and our LaFIn, respectively. }
		\vspace{-3pt}
		\label{quanlitative comparison}
\end{figure*}

\subsection{Training Strategy}
The generator is desired to complete image via $\hat{I}\defeq\mc{G}(I^M)$. For face images, their strong regularity, like the landmarks considered by our design $\hat{I}\defeq\mc{G}_P(\mc{G}_L(I^M), I^M)$, could benefit model reduction and training procedure, as the space is considerably restricted by the regularity. Intuitively, the training for $\mc{G}_P$ and $\mc{G}_L$ can be finished jointly. Technically, it is feasible. But, in practice, it is not a good choice. The reasons are as follows: 1) the loss for $\mc{G}_L$, say $\mathcal{L}_{lmk}$, computes over a small number of (only $68$ in this work) \textbf{locations}, which is incompatible with  $\mathcal{L}_{inp}$. In other words, the parameter tuning is extremely hard; and 2) even with the well-tuned parameters, the performance of both $\mc{G}_L$ and $\mc{G}_P$ may be too inaccurate especially at the beginning of training, which consequently leads to low-quality landmark prediction and inpainting results. These two coupled factors very likely drag the training into dilemmas, like bad points of convergence and/or high prices of training.  Thus, we decouple the joint model into the landmark prediction and inpainting modules, and train them separately. It is worth to note that we actually have trained the model in a joint way with different carefully-tuned settings, the best shot is still inferior to our separate training. In experiments shown in this work, the landmark prediction model and the inpainting model are trained using $256\times256$ images and optimized by the Adam optimizer \cite{kingma2014adam} with $\beta_1=0$ and $\beta_2=0.9$, and the learning rate $=10^{-4}$. The learning rate of the discriminator is $10^{-5}.$  We use batch size $= 16$ for the landmark prediction module and batch size $= 4$ for the inpainting model.

\section{Experimental Validation on Face Inpainting}

\begin{figure*}
		\begin{center}
		\subfigure[Ground truth (GT)]{
			\begin{minipage}[t]{0.15\linewidth}
				\includegraphics[width=1.13in]{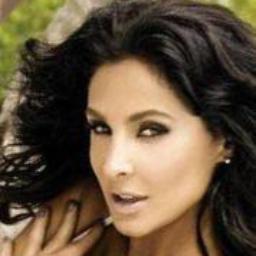}
				\includegraphics[width=1.13in]{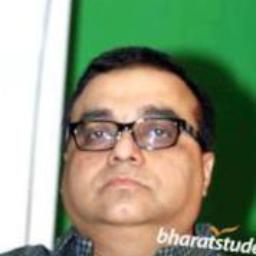}				
			\end{minipage}
		}
		\subfigure[Masked GT]{
			\begin{minipage}[t]{0.15\linewidth}
				\includegraphics[width=1.13in]{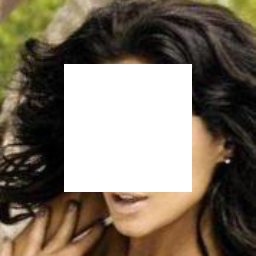}
				\includegraphics[width=1.13in]{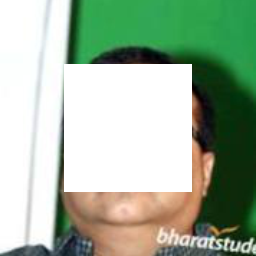}
			\end{minipage}
		}
		\subfigure[CE]{
			\begin{minipage}[t]{0.15\linewidth}
				\includegraphics[width=1.13in]{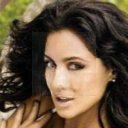}
				\includegraphics[width=1.13in]{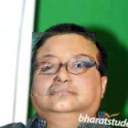}
			\end{minipage}
		}
		\subfigure[GFC]{
			\begin{minipage}[t]{0.15\linewidth}
				\includegraphics[width=1.13in]{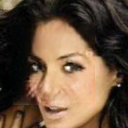}
				\includegraphics[width=1.13in]{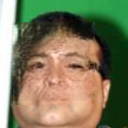}
			\end{minipage}
		}
		\subfigure[EC]{
			\begin{minipage}[t]{0.15\linewidth}
				\includegraphics[width=1.13in]{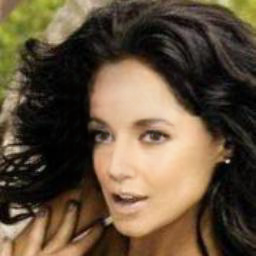}
				\includegraphics[width=1.13in]{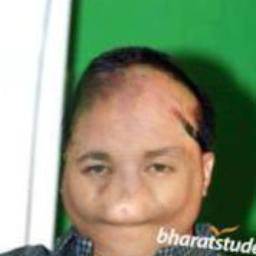}
			\end{minipage}
		}
		\subfigure[Ours]{
			\begin{minipage}[t]{0.15\linewidth}
				\includegraphics[width=1.13in]{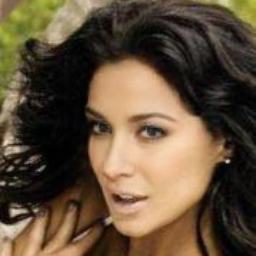}
				\includegraphics[width=1.13in]{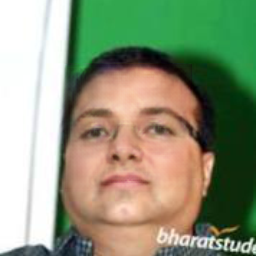}
			\end{minipage}
		}
		
	\end{center}
	\vspace{-10pt}
	\caption{Visual comparison between the competitors on the CelebA dataset. (a) shows the ground-truth images. (b) depicts the masked versions of (a). (c)-(f) are the results obtained by CE, GFC, EC, and our LaFIn, respectively.}
		\vspace{-10pt}
	\label{fig:canew}
\end{figure*}

\begin{figure*}
		\begin{center}
		\subfigure[Ground truth (GT)]{
			\begin{minipage}[t]{0.15\linewidth}
								\includegraphics[width=1.13in]{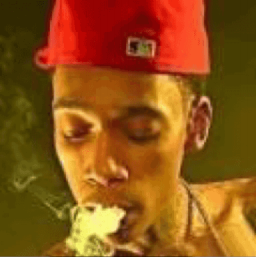}
								\includegraphics[width=1.13in]{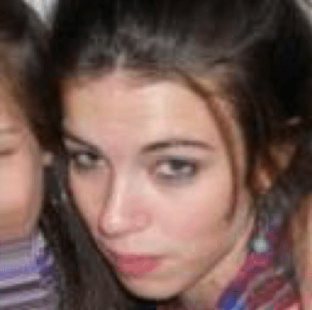}
			\end{minipage}
		}
		\subfigure[Masked GT]{
			\begin{minipage}[t]{0.15\linewidth}
								\includegraphics[width=1.13in]{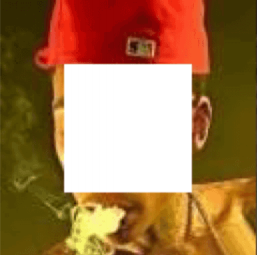}
								\includegraphics[width=1.13in]{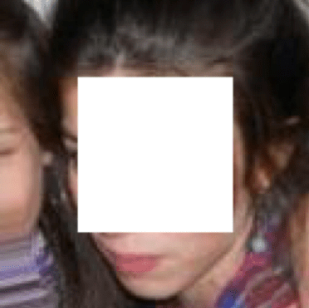}
			\end{minipage}
		}
		\subfigure[CE]{
			\begin{minipage}[t]{0.15\linewidth}
				\includegraphics[width=1.13in]{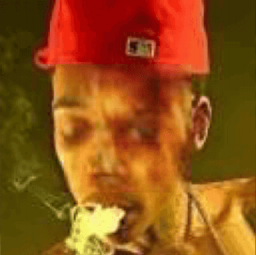}
				\includegraphics[width=1.13in]{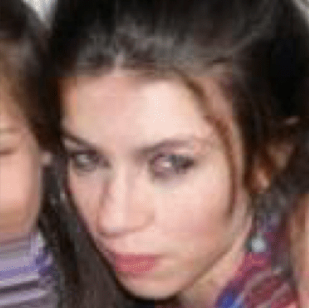}
				
			\end{minipage}
		}
		\subfigure[GFC]{
			\begin{minipage}[t]{0.15\linewidth}
								\includegraphics[width=1.13in]{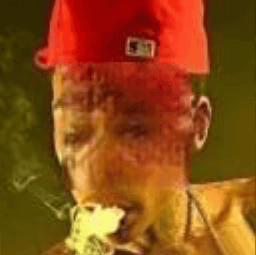}
								\includegraphics[width=1.13in]{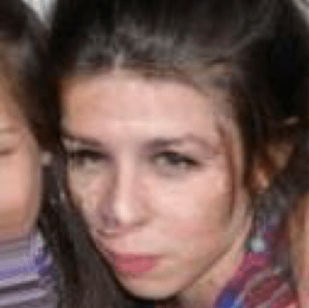}
			\end{minipage}
		}
		\subfigure[GAFC]{
			\begin{minipage}[t]{0.15\linewidth}
								\includegraphics[width=1.13in]{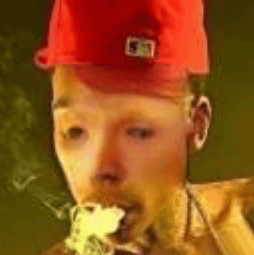}
								\includegraphics[width=1.13in]{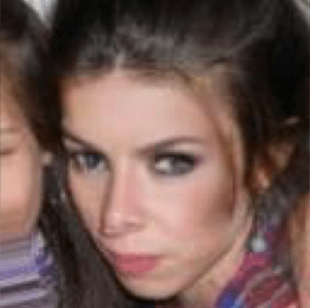}

			\end{minipage}
		}
		\subfigure[Ours]{
			\begin{minipage}[t]{0.15\linewidth}
								\includegraphics[width=1.13in]{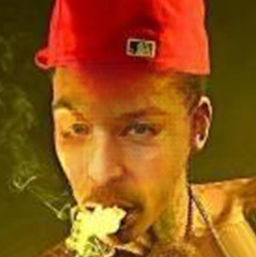}
								\includegraphics[width=1.13in]{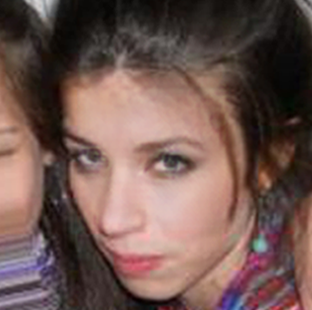}
			\end{minipage}
		}
		
	\end{center}
	\vspace{-17pt}
	\caption{Visual comparison between the competitors on the CelebA dataset. (a) shows the ground-truth images. (b) depicts the masked versions of (a). (c)-(f) are the results obtained by CE, GFC, GAFC, and our LaFIn, respectively.}
		\vspace{-14pt}
	\label{fig:ca}
\end{figure*}

\begin{table}[t]
	\begin{center}
		\begin{tabular}{c|c|c|c|c|c}
			\multicolumn{2}{r|}{{\bf Mask}}&CA&PIC&EC& Ours\\
			\hline
			\multirow{6}*{\rotatebox{90}{PSNR}}&10-20\%&27.51&30.33&30.73&{\bf 31.48}\\
			\cline{2-6}
			&20-30\%&24.42&27.05&27.56&{\bf 28.31}\\
			\cline{2-6}
			&30-40\%&22.14&24.68&25.34&{\bf 26.14}\\
			\cline{2-6}
			&40-50\%&20.29&22.58&23.44&{\bf 24.22}\\
			\cline{2-6}
			&50\%+&18.10&19.54&20.71&{\bf 21.61}\\
			\cline{2-6}
			&Center&24.13&24.22&24.82&{\bf 25.92}\\
			\hline
			\multirow{6}*{\rotatebox{90}{SSIM}}&10-20\%&0.942&0.968&0.971&{\bf 0.975}\\
			\cline{2-6}
			&20-30\%&0.892&0.936&0.942&{\bf 0.951}\\
			\cline{2-6}
			&30-40\%&0.832&0.894&0.907&{\bf 0.922}\\
			\cline{2-6}
			&40-50\%&0.761&0.838&0.859&{\bf 0.883}\\
			\cline{2-6}
			&50\%+&0.646&0.715&0.754&{\bf 0.805}\\
			\cline{2-6}
			&Center&0.864&0.870&0.874&{\bf 0.905}\\
			\hline
			\multirow{6}*{\rotatebox{90}{FID}}&10-20\%&7.29&2.72&2.33&{\bf 2.05}\\
			\cline{2-6}
			&20-30\%&14.62&4.49&3.89&{\bf 3.36}\\
			\cline{2-6}
			&30-40\%&24.41&6.56&5.94&{\bf 5.11}\\
			\cline{2-6}
			&40-50\%&38.83&9.24&8.78&{\bf 7.12}\\
			\cline{2-6}
			&50\%+&51.21&13.32&13.92&{\bf 10.84}\\
			\cline{2-6}
			&Center&7.39&{\bf 4.98}&8.26&6.63\\
			\hline
		\end{tabular}
	\end{center}
	\vspace{-5pt}
	\caption{Quantitative comparison on the CelebA-HQ dataset in terms of PSNR, SSIM and FIC on random and center masks. }
		\vspace{-5pt}
		\label{quantative1}
\end{table}

\begin{table}[t]
	\begin{center}
		\begin{tabular}{|c|c|c|c|c|}
		    \hline
			{\bf Metric}&CE&GFC&EC&Ours\\
			\hline
			PSNR&25.46&21.04&25.83&{\bf 26.25}\\
			\hline
			SSIM&0.909&0.766&0.899&{\bf 0.912}\\
			\hline
			FID&{\bf 1.731}&14.958&3.519& 3.512\\
			\hline
		\end{tabular}
	\end{center}
	\vspace{-5pt}
	\caption{Quantitative comparison on the CelebA dataset in PSNR, SSIM and FID on center masks. }
		\vspace{-7pt}
		\label{quantative0}
\end{table}

In this part, we evaluate the face inpainting performance of our LaFIn on the CelebA-HQ face dataset \cite{liu2015deep,karras2018progressive}. The masks used for training come from the random mask dataset \cite{liu2018image} and additional block masks randomly generated. The competitors involved in the comparison include Context Encoder (CE) \cite{CE}, Generative Face Completion (GFC) \cite{GFC}, Contextual Attention (CA) \cite{yu2018generative}, Geometry Aware Face Completion (GAFC) \cite{GAFC}, Pluralistic Image Completion (PIC)  \cite{zheng2019pluralistic}, and EdgeConnect (EC) \cite{nazeri2019edgeconnect}.
For quantitatively measuring the performance difference among the competitors, we employ PSNR, SSIM  \cite{wang2004image} and FID \cite{heusel2017gans}, as metrics. For PSNR and SSIM, higher values indicate better performance, while for FID, the lower the better. As the ground-truth landmarks are unavailable for the CelebA-HQ dataset, we apply the results by FAN \cite{bulat2017far} to perform as the reference information for training our landmark predictor. 



\begin{figure}[t]
	\subfigure[]{
		\begin{minipage}[t]{0.17\linewidth}
			\includegraphics[width=0.75in]{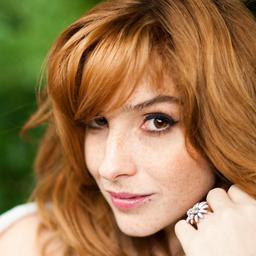}
			\includegraphics[width=0.75in]{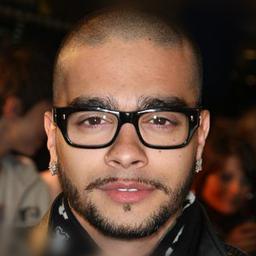}
			\includegraphics[width=0.75in]{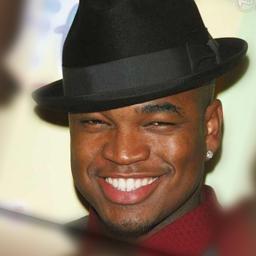}
			\includegraphics[width=0.75in]{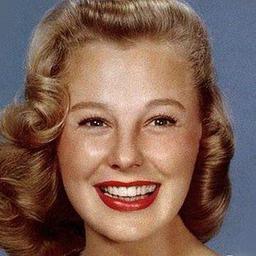}
		\end{minipage}
	}
	\hspace{.07in}
	\subfigure[]{
		\begin{minipage}[t]{0.17\linewidth}
			\includegraphics[width=0.75in]{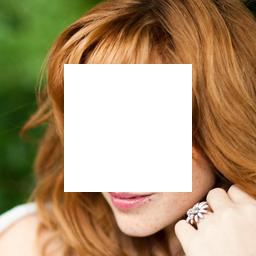}
			\includegraphics[width=0.75in]{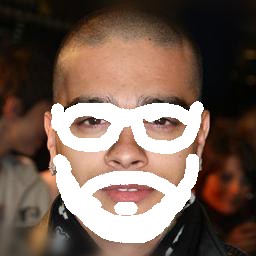}
			\includegraphics[width=0.75in]{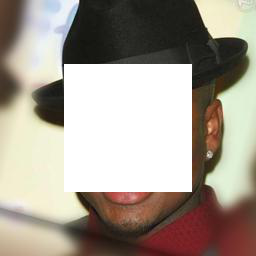}
			\includegraphics[width=0.75in]{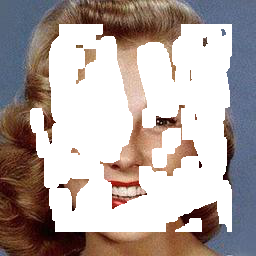}
		\end{minipage}
	}
	\hspace{.07in}
	\subfigure[]{
		\begin{minipage}[t]{0.17\linewidth}
			\includegraphics[width=0.75in]{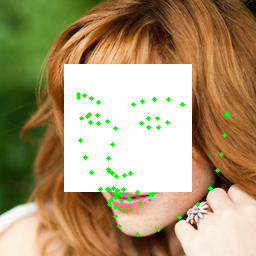}
			\includegraphics[width=0.75in]{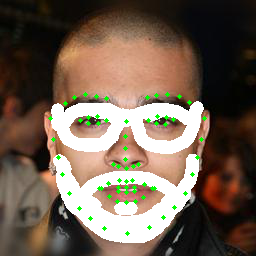}
			\includegraphics[width=0.75in]{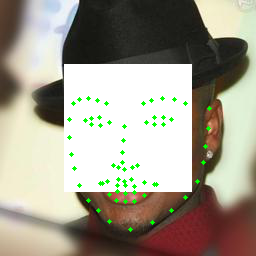}
			\includegraphics[width=0.75in]{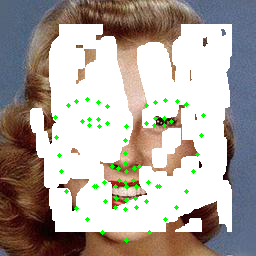}
		\end{minipage}
	}
	\hspace{.07in}
	\subfigure[]{
		\begin{minipage}[t]{0.17\linewidth}
			\includegraphics[width=0.75in]{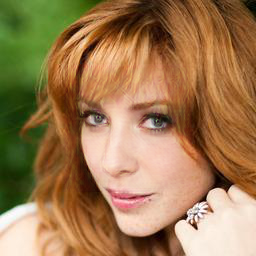}
			\includegraphics[width=0.75in]{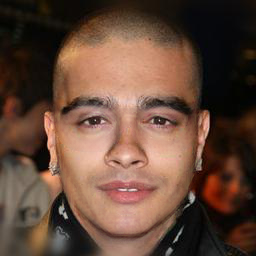}
			\includegraphics[width=0.75in]{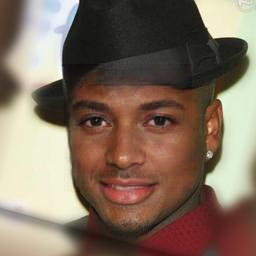}
			\includegraphics[width=0.75in]{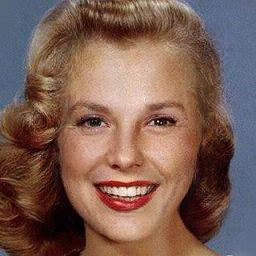}
		\end{minipage}
	}
	\vspace{-10pt}
	\caption{More results by LaFIn. (a) Ground truth images. (b) Masked images. (c) Predicted landmarks. (d) Results.}
		\vspace{-14pt}
	\label{fig:more}
\end{figure}


\begin{table}[t]
	\begin{center}
			\begin{tabular}{c|c|c|c|c}
				\multicolumn{2}{r|}{{\bf Mask}}&\makecell[c]{w/o LSTA}&\makecell[c]{w/o LMK}&\makecell[c]{Ours}\\
				\hline
				\multirow{3}*{\rotatebox{90}{PSNR}}&10-20\%&31.02&31.10&{\bf 31.48}\\
				\cline{2-5}
				&40-50\%&24.00 &23.75&{\bf 24.22}\\
				\cline{2-5}
				&Center&25.75&24.90&{\bf 25.92}\\
				\hline
				\multirow{3}*{\rotatebox{90}{SSIM}}&10-20\%&0.973&0.972&{\bf 0.9754}\\
				\cline{2-5}
				&40-50\%& 0.879 &0.869 &{\bf 0.883}\\
				\cline{2-5}
				&Center&0.902&0.879&{\bf 0.905}\\
				\hline
				\multirow{3}*{\rotatebox{90}{FID}}&10-20\%&2.18&2.26&{\bf 2.05}\\
				\cline{2-5}
				&40-50\%& 8.25 &7.72&{\bf 7.12}\\
				\cline{2-5}
				&Center&8.56&7.21&{\bf 6.63}\\
				\hline
			\end{tabular}
	\end{center}
	\vspace{-5pt}
	\caption{Ablation study on different configurations of LaFIn.}
	\vspace{-7pt}
	\label{tab:as}
\end{table}

\textit{Result comparison.}  
Table \ref{quantative1} reports the performance of CA, EC, PIC and our LaFIn with different types and sizes of mask. Notice that for CA and PIC, the pre-trained models on the CelebA-HQ are given\footnote{\scriptsize{\url{https://github.com/JiahuiYu/generative_inpainting}}}\footnote{\scriptsize{\url{https://github.com/lyndonzheng/Pluralistic-Inpainting}}}. While the authors of EC do not offer the pre-trained model on the CelebA-HQ dataset, we try our best to retrain it using the training code\footnote{\scriptsize{\url{https://github.com/knazeri/edge-connect}}}.  As can be seen from the numbers in Table \ref{quantative1}, EC is superior over PIC and CA in most cases, as it employs the edge information to help inpainting. Overall, our LaFIn outperforms the others by large margins in terms of all PSNR, SSIM and FID, except for the case of center falling behind PIC in terms of FID $6.63$ \textit{vs.} $4.98$, the explanation is in Appendix. This comparison verifies that the landmarks are stronger and more robust guidance than the edges for the task of face inpainting. Further quantitative comparisons with CE, EC and GFC under center masks on CelebA are shown in Table \ref{quantative0}. Figure \ref{quanlitative comparison} depicts two visual comparisons among CA, EC, PIC, and our LaFIn,  from which, we can see that LaFIn can generate more natural-looking and visually striking results even on the cases with large poses and extreme occlusions.  Figure \ref{fig:canew} and \ref{fig:ca}  further provide visual comparisons of CE, GFC, GAFC, EC and LaFIn on four samples from the CelebA dataset. Notice that GFC utilizes the face parsing information and GAFC uses both the landmark and parsing to guide the inpainting. As observed from the results, those by GFC suffer from the face component shifting problem. GAFC\footnote{Since neither the code nor implementation details of GAFC is available, when this paper is prepared, we only compare the cases cropped from the GAFC paper.} seems to somewhat mitigate the problem due to the introduction of landmarks, but still inferior to our LaFIn. This comparison tells that the redundancy of the face parsing prior may alternatively hurt the performance. It is worth to emphasize that GAFC considers the symmetry property of faces and low rankness of mask in the loss, which are not so reasonable because  large poses of faces and random corruptions can easily violate these properties. Also, the editing on parsed regions (+landmarks) is much more difficult than on sparse landmarks. Figure \ref{fig:more} gives several more results by LaFIn. Due to page limit, please find more comparisons in Appendix.

\begin{figure}[t]
	\begin{center}
		\subfigure[]{
				\includegraphics[width=0.23\linewidth]{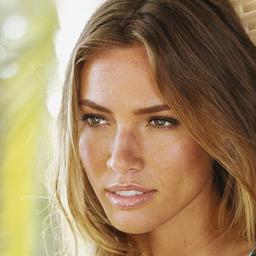}
				
		}\hspace{-5pt}
		\subfigure[]{
				\includegraphics[width=0.23\linewidth]{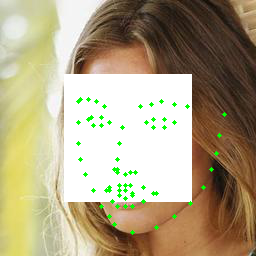}
		}\hspace{-5pt}
		\subfigure[]{

					\includegraphics[width=0.23\linewidth]{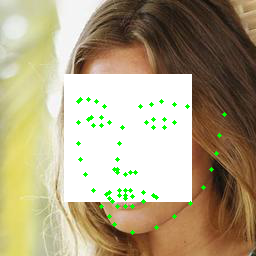}
		}\hspace{-5pt}
		\subfigure[]{
			\includegraphics[width=0.23\linewidth]{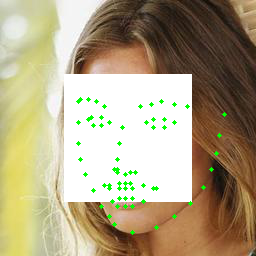}
		}\\
		\subfigure[]{

					\includegraphics[width=0.23\linewidth]{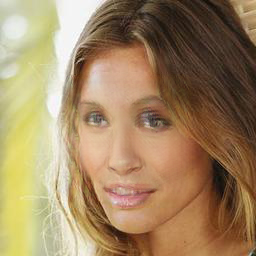}
		}\hspace{-5pt}
		\subfigure[]{
				\includegraphics[width=0.23\linewidth]{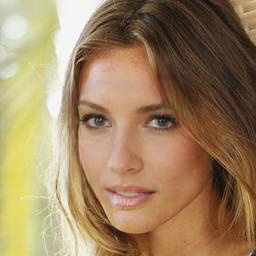}
		}\hspace{-5pt}
		\subfigure[]{
						\includegraphics[width=0.23\linewidth]{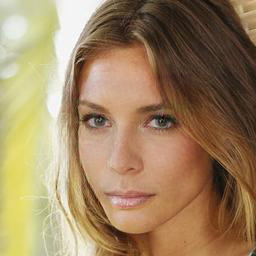}
				
		}\hspace{-5pt}
		\subfigure[]{
			\includegraphics[width=0.23\linewidth]{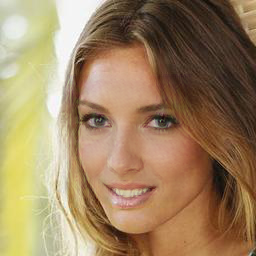}
		}
	\end{center}
\vspace{-10pt}
	\caption{(a) Ground truth image. (b),(c) and (d) present three versions of landmark on the masked image. (f), (g) and (h) give the results conditioned on (b), (c), and (d), respectively. (e) the result without landmark information.}
	\vspace{-7pt}\label{fig:lmk}
\end{figure}

\textit{Ablation study.} Table \ref{tab:as} reports the difference of LaFIn with the long short term attention (LSTA) disabled (denoted as w/o LSTA), LaFIn with the landmark guidance canceled (w/o LMK) , and the complete LaFIn.  From the numbers reported in Table \ref{tab:as}, both the LSTA and LMK help the task of face inpainting. Specifically, the LSTA influences more than the landmark indication on the cases with relatively small corruptions. This phenomenon is reasonable because the completed part should pay more attention on the attribute consistency to make the results visually coincident to the observed (large) area. While for the cases with relatively large masks, the attribute consistency is barely violated in the generated result as there is few information given to match. Alternatively the landmark information is more important to ensure the structure well-preserved. The above corroborates the principle of our design, say the LSTA is for the attribute consistency and the landmarks for the main structure.
To view the effect of landmark, Figure \ref{fig:lmk} shows the inpainting results based on different landmark templates. By varying the templates (mouth), the completed faces accordingly change with much better visual quality than the one without adopting any landmark information. This experiment also informs us that editing faces is viable by manipulating the landmark template. Affirmatively, operating sparse landmarks is more convenient than modifying parsed regions (together with landmarks \cite{GAFC}).



\section{Further Finding on Data Augmentation}
\begin{figure}[t]
	\centering
	\includegraphics[width=0.23\linewidth]{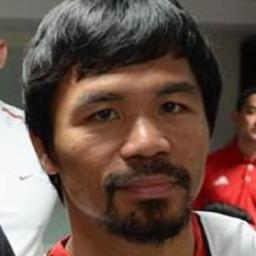}
	\includegraphics[width=0.23\linewidth]{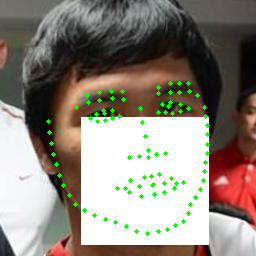}
	\includegraphics[width=0.23\linewidth]{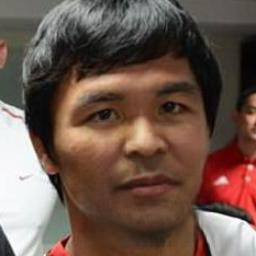}
	\includegraphics[width=0.23\linewidth]{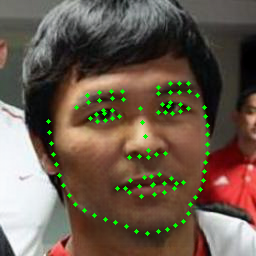}
	\caption{ From left to right: the original image, masked image with ground-truth landmarks, augmented result, and augmented image with ground-truth landmarks.}
	\vspace{-5pt}
	\label{aug_figure}
\end{figure}

Most of data-driven approaches, if not all, require well-labeled data, which is time consuming and labor intensive. Like the original motivation of GANs, it attempts to produce more samples for training networks. Specifically for facial landmark detectors/predictors, one wants to generate diverse plausible faces given the ground-truth landmarks. Intrinsically, this is how our work stands. For an image $I$, we are able to obtain the augmented data $I_{aug}$ through $I_{aug} \defeq \mc{G}_{P}(L_{gt}, M\circ I)$, where $L_{gt}$ is the landmark of $I$, and $M$ stands for any mask. By doing so, for the image $I$, the augmented faces vary with different masks. The discriminator will make sure that the inpainted results match $L_{gt}$.  An example is shown in Fig. \ref{aug_figure}, from which we can see that the features of $I_{aug}$ are significantly different from those of $I$ with the same landmarks. Consequently, the pair of ($I_{aug}$, $L_{gt}$) can be used for training.


\begin{table}[t]
	\begin{center}
		\begin{tabular}{c|c|c|c}
			\hline
			LAB & LAB$_{aug}$& LaFIn & LaFIn$_{aug}$\\
			\hline
			\hline
			5.66 & 5.43& 6.79 & 5.92\\
			\hline
		\end{tabular}
	\end{center}
\vspace{-10pt}
	\caption{Comparison in NME on the WFLW dataset.}
	\vspace{-5pt}
	\label{tab:wflw}
	
		\begin{center}
			\begin{tabular}{c|ccc}
				\hline
				Dataset&\makecell[c]{Common}&\makecell[c]{Challenging}&Full\\
				\hline
				\hline
				LaFIn &{ 4.69}&{ 8.95}&{ 5.42}\\
				LaFIn$_{aug}$&{ 4.45}&{ 8.91}&{ 5.21}\\
				\hline
			\end{tabular}
		\end{center}
	\vspace{-12pt}
		\caption{Comparison in NME on the 300W dataset.}
		\vspace{-13pt}
\end{table}

To validate the effectiveness of such a data-augmentation manner, we feed the augmented data into both our $\mc{G}_L$ and LAB \cite{LAB} on the WFLW dataset \cite{LAB}.  We notice that LAB\footnote{We use the PyTorch version of LAB downloaded from\\ \scriptsize{\url{https://github.com/FunkyKoki/Look_At_Boundary_PyTorch}}} is carefully built for the task of facial landmark detection, while the landmark module in LaFIn is much simpler and smaller because as previously explained, in our task, the landmarks can be not that accurate as long as they can provide the main structure of faces. Therefore, our performance in NME (normalized mean error by inter-ocular factor) is inferior to LAB. Nevertheless,  as can be viewed from Table \ref{tab:wflw}, the augmentation improves both the performance of LAB and our LaFIn. In addition, we also test LaFIn on the 300W dataset, the numerical results consistently reveal the effectiveness of the augmentation. Notice that no obvious difference in inpainted results is observed using the landmark predictors without and with augmentation, which again verifies that our inpainting module is robust against variation in landmarks, and can produce striking results as long as the structure is reasonably offered.

\section{Conclusion}
In this study, we have developed a generative network, namely LaFIn, for completing face images. The proposed LaFIn first predicts the landmarks then accomplishes the inpainting conditioned on the predicted landmarks. Our principle is that the landmarks are neat, sufficient, and robust to perform as guidance for providing the structural information to the face inpainting module. For ensuring the attribute consistency, we designed to harness distant spatial context and connect temporal feature maps. Extensive experiments have been conducted to verify our claims, reveal the efficacy of our design and, demonstrate its advances over state-of-the-art alternatives both qualitatively and quantitatively. 
Furthermore, we proposed to use our LaFIn to augment face-landmark data for relieving manual annotation in the task of landmark detection. The effectiveness of this manner has been experimentally confirmed.

\appendix
\appendixpage
\begin{appendices}

\section{Network Architecture} 
\subsection{The Landmark Predictor}
Our landmark predictor is based on the MobileNet-V2. A series of bottlenecks are employed to extract the features and speed up the network. Feature maps at different stages of fusion layers are fully connected to achieve the final landmark prediction. The detailed architecture is shown in Table \ref{tab:lp_arch}.

\begin{table}[h]
\begin{center}
\begin{tabular}{c|c|c|c|c|c}
\hline
\textbf{Input}&\textbf{Operator}&$t$&$c$&$n$&$s$\\
\hline\hline
$256^2\times3$&conv2d&-&32&1&2\\
$128^2\times32$&bottleneck&1&16&1&1\\
$128^2\times16$&bottleneck&6&24&2&2\\
$64^2\times24$&bottleneck&6&32&3&2\\
$32^2\times32$&bottleneck&6&64&4&2\\
$16^2\times64$&bottleneck&6&96&3&1\\
$16^2\times96$&bottleneck&6&160&3&2\\
$8^2\times160$&bottleneck&6&320&1&1\\
(C1) $8^2\times320$&conv2d 1x1&-&1280&1&1\\
(C2) $8^2\times1280$&avgpool 8x8&-&-&1&-\\
$1\times1\times1280$&conv2d 1x1&-&64&-&-\\
(S3) $1\times1\times64$&-&-&64&-&-\\
\hline
C1&conv2d 1x1&-&128&1&1\\
$8^2\times128$&avgpool 8x8&-&128&1&-\\
(S1) $1\times1\times128$&-&-&128&1&-\\
\hline
C2&conv2d 1x1&-&128&1&1\\
$8^2\times128$&avgpool 8x8&-&128&1&-\\
(S2) $1\times1\times128$&-&-&128&1&-\\
\hline
S1,S2,S3&Full Connection&-&136&1&-\\
\hline
\end{tabular}
\end{center}
\caption{The network architecture of our landmark predictor. Each line represents a sequence of identical layers, repeating $n$ times. For layers in the same sequence, they have the same number $c$ of output channels. The first convolution layer of each sequence has a stride $s$. The expansion factor is applied to the input size in bottleneck layers.}
\label{tab:lp_arch}
\end{table}

\subsection{Discriminator of the Inpaintor}
The discriminator is built upon the Patch-GAN architecture. Spectral normalization is applied on the convolution layers to stabilize the training process. The attention block is placed in the discriminator to adaptively treat the features. The detailed architecture is given in Table \ref{tab:dis_arch}.
\begin{table}[t]
\begin{center}
\begin{tabular}{c|c|c|c|c|c}
\hline
\textbf{Input}&\textbf{Operator}&$k$&$c$&$s$&$p$\\
\hline\hline
$256^2\times4$&Conv-SN-LReLU&4&64&2&1\\
$128^2\times64$&Conv-SN-LReLU&4&128&2&1\\
$128^2\times128$&Attention &-&128&-&-\\
$64^2\times128$&Conv-SN-LReLU&4&256&2&1\\
$32^2\times256$&Conv-SN-LReLU&4&512&1&1\\
$31^2\times512$&Conv-SN-Sigmoid&4&1&1&1\\
$30^2\times1$&-&-&1&-&-\\
\hline
\end{tabular}
\end{center}
\caption{The discriminator network architecture. Each line represents a sequence of listed layers or a hole block. The $k,c,s,p$ represent kernel size, output channels, stride and padding of convolution or deconvolution layers, respectively. SN refers to spectral normalization and LReLU means leaky relu with the slope set to $0.2$.}
\label{tab:dis_arch} 
\end{table}

\subsection{Generator of the Inpaintor}
The generator is based on the U-Net structure. Three encoding blocks are applied for down-sampling, followed by 7 residual blocks with dilated convolutions to enlarge receptive fields. The long-short term attention block connects the features from the last residual block and the last down-sampling block so that the features in a wider range can be better used. The shortcuts are added between corresponding encoder and decoders. The 1x1 convolutions are employed as channel attention to adjust the weights of features from shortcut and last layer. The detailed architecture is shown in Table \ref{tab:gen_arch}.

\begin{table*}[ht]
\begin{center}
\begin{tabular}{c|c|c|c|c|c|c}
\hline
\textbf{Input}&\textbf{Operator}&$k$&$c$&$s$&$p$&\textbf{Out}\\
\hline\hline
$256^2\times4$&Conv-IN-ReLU&7&64&1&3&E1\\
$256^2\times64$&Conv-IN-ReLU&4&128&2&1&E2\\
$128^2\times128$&Conv-IN-ReLU&4&256&2&1&E3\\
$64^2\times256$&Dilated Residual Block&-&256&-&-&-\\
$64^2\times256$&Dilated Residual Block&-&256&-&-&-\\
$64^2\times256$&Dilated Residual Block&-&256&-&-&-\\
$64^2\times256$&Dilated Residual Block&-&256&-&-&-\\
$64^2\times256$&Dilated Residual Block&-&256&-&-&-\\
$64^2\times256$&Dilated Residual Block&-&256&-&-&-\\
$64^2\times256$&Dilated Residual Block&-&256&-&-&R7\\
E3,R7 $64^2\times512$&Short+Long Term Attention&-&256&-&-\\ $64^2\times256$&Deconv-IN-ReLU&4&128&2&1&D1\\
E2,D1 $128^2\times256$&Conv-IN-ReLU&1&256&1&0&-\\
$128^2\times256$&Deconv-IN-ReLU&4&64&2&1&D2\\
E1,D2 $256^2\times128$&Conv-IN-ReLU&1&128&1&0&-\\
$256^2\times128$&Conv-IN-tanh&7&3&1&3&-\\
$256^2\times3$&-&-&3&-&-&-\\
\hline
\end{tabular}
\end{center}
\caption{The generator network architecture. Each line represents a sequence of listed layers or a hole block. The $k,c,s,p$ represent kernel size, output channels, stride and padding of convolution or deconvolution layers, respectively. Reflection paddings are applied in the first convolution layer and last deconvolution layer while others apply zero-padding. The layers with skip connections are showed at first column. Other layers directly take output of previous layers as input. IN represents instance normalization.} 
\label{tab:gen_arch}
\end{table*}

\section{Implementation Details on Data Augmentation}
In the last but one section of the main paper, we validated the effectiveness of the proposed data-augmentation manner. The implementation details are as follows. In our experiment, for each pair of training sample$(I_{gt},L_{gt})$ in a single epoch, a pair of augmented data $(I_{aug},L_{gt})$ will be generated by the inpaintor and be applied as the additional training data. Moreover, in different epochs, the masked region will change so that various augmented images can be produced. The training settings of LaFIn is same as above mentioned except the batch size shrinks to 4. The settings of LAB\cite{LAB} follow its original implementation.

\section{Further Analysis on Experimental Results}
\begin{figure*}[ht]
	\begin{center}
		\subfigure[Ground truth (GT)]{
			\begin{minipage}[t]{0.15\linewidth}
				\includegraphics[width=1.13in]{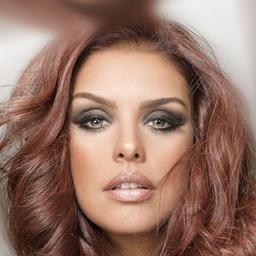}
				\includegraphics[width=1.13in]{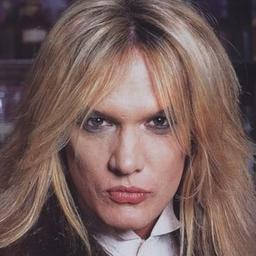}
				\includegraphics[width=1.13in]{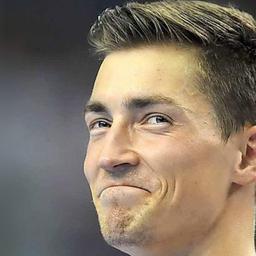}
				\includegraphics[width=1.13in]{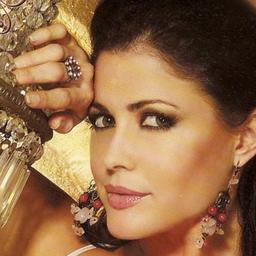}
				\includegraphics[width=1.13in]{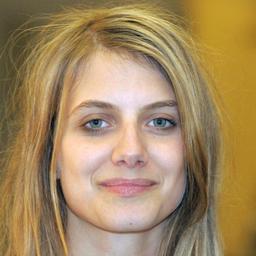}
				\includegraphics[width=1.13in]{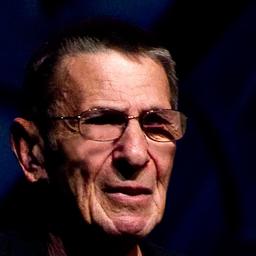}
				\includegraphics[width=1.13in]{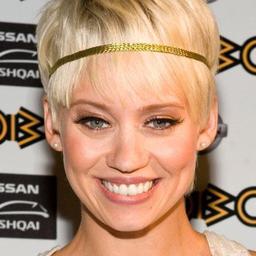}

			\end{minipage}
		}
		\subfigure[Masked GT]{
			\begin{minipage}[t]{0.15\linewidth}
				\includegraphics[width=1.13in]{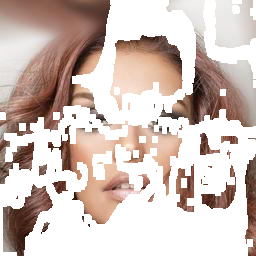}
				\includegraphics[width=1.13in]{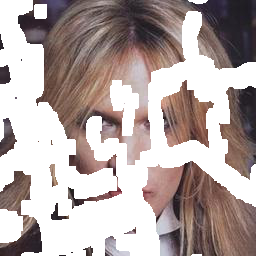}
				\includegraphics[width=1.13in]{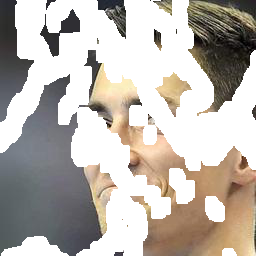}
				\includegraphics[width=1.13in]{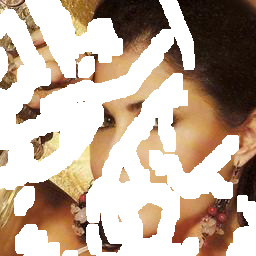}
				\includegraphics[width=1.13in]{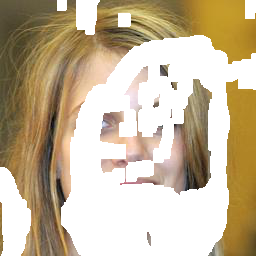}
				\includegraphics[width=1.13in]{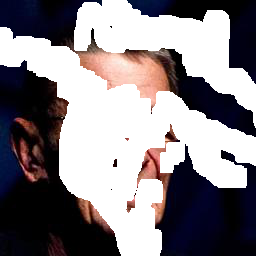}
				\includegraphics[width=1.13in]{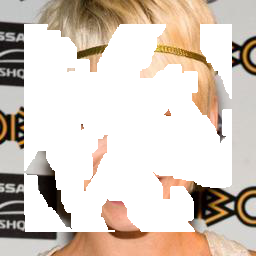}
			\end{minipage}
		}
		\subfigure[CA]{
			\begin{minipage}[t]{0.15\linewidth}
				\includegraphics[width=1.13in]{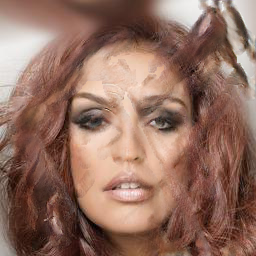}
				\includegraphics[width=1.13in]{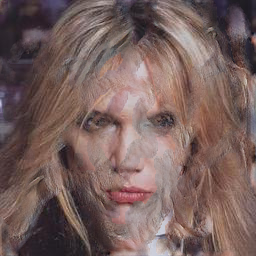}
				\includegraphics[width=1.13in]{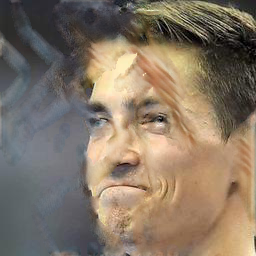}
				\includegraphics[width=1.13in]{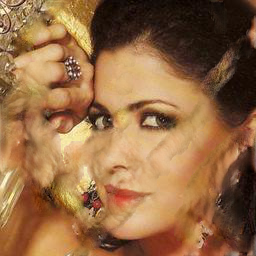}
				\includegraphics[width=1.13in]{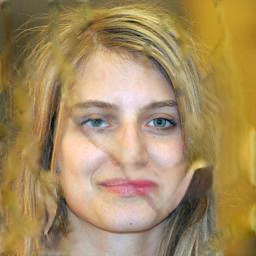}
				\includegraphics[width=1.13in]{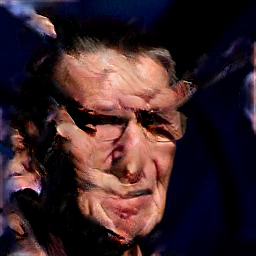}
				\includegraphics[width=1.13in]{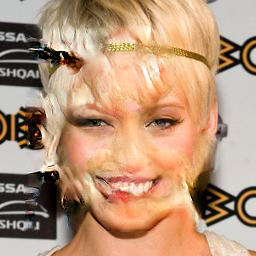}
			\end{minipage}
		}
		\subfigure[EC]{
			\begin{minipage}[t]{0.15\linewidth}
				\includegraphics[width=1.13in]{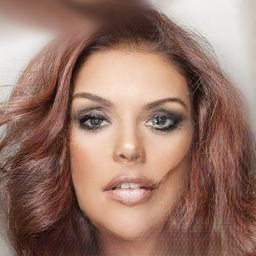}
				\includegraphics[width=1.13in]{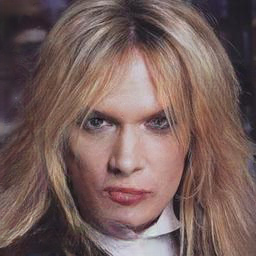}
				\includegraphics[width=1.13in]{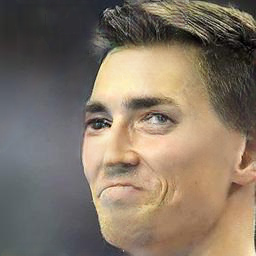}
				\includegraphics[width=1.13in]{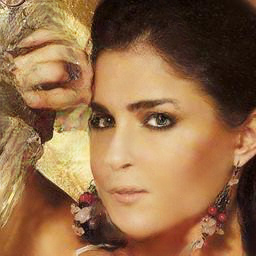}
				\includegraphics[width=1.13in]{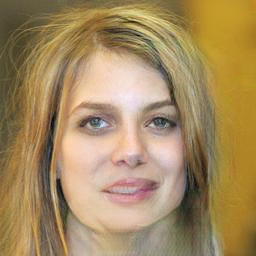}
				\includegraphics[width=1.13in]{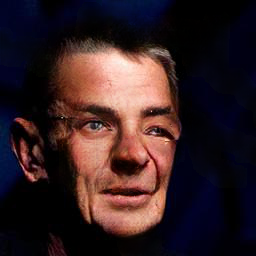}
				\includegraphics[width=1.13in]{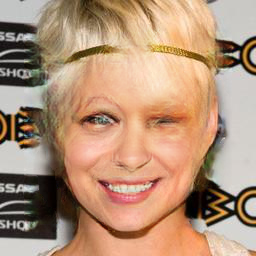}

			\end{minipage}
		}
		\subfigure[PIC]{
			\begin{minipage}[t]{0.15\linewidth}
				\includegraphics[width=1.13in]{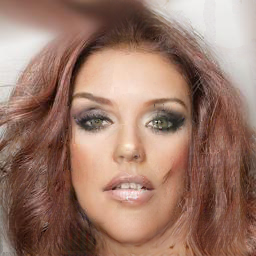}
				\includegraphics[width=1.13in]{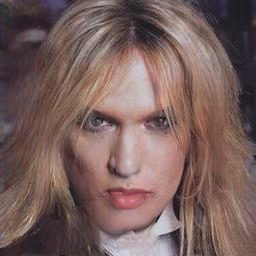}
				\includegraphics[width=1.13in]{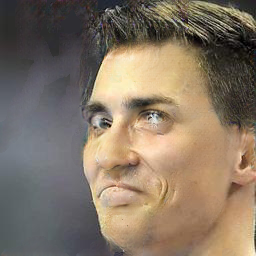}
				\includegraphics[width=1.13in]{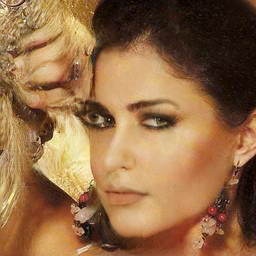}
				\includegraphics[width=1.13in]{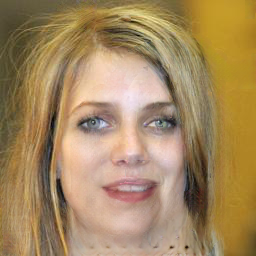}
				\includegraphics[width=1.13in]{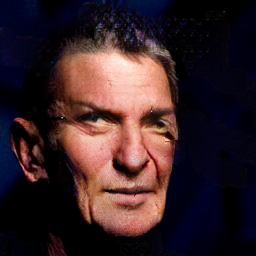}
				\includegraphics[width=1.13in]{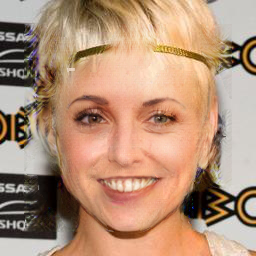}
			\end{minipage}
		}
		\subfigure[Ours]{
			\begin{minipage}[t]{0.15\linewidth}
				\includegraphics[width=1.13in]{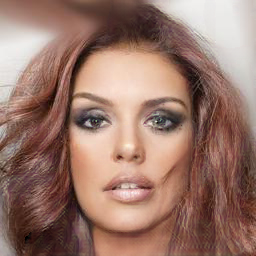}
				\includegraphics[width=1.13in]{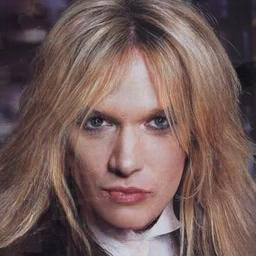}
				\includegraphics[width=1.13in]{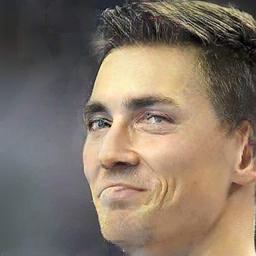}
				\includegraphics[width=1.13in]{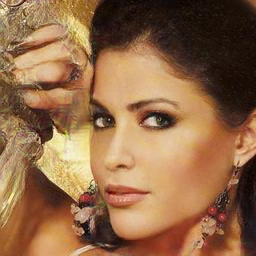}
				\includegraphics[width=1.13in]{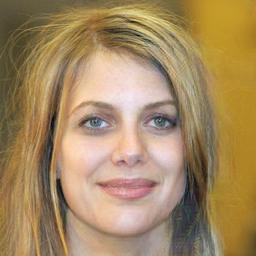}
				\includegraphics[width=1.13in]{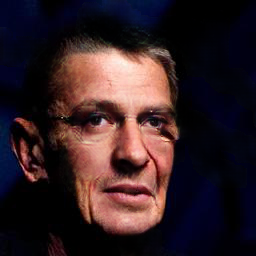}
				\includegraphics[width=1.13in]{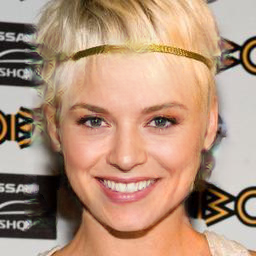}
			\end{minipage}
		}
	\end{center}
\vspace{-7pt}
	\caption{More results with other state-of-the-art techniques on the CelebA-HQ dataset. (a) shows the ground-truth images. (b) depicts the masked versions of (a). (c)-(f) are the results obtained by CA, EC, PIC, and our LaFIn, respectively. }
		\vspace{-7pt}
		\label{fig:more1}
\end{figure*}

\begin{figure*}[ht]
	\begin{center}
		\subfigure[Ground truth (GT)]{
			\begin{minipage}[t]{0.15\linewidth}
				\includegraphics[width=1.13in]{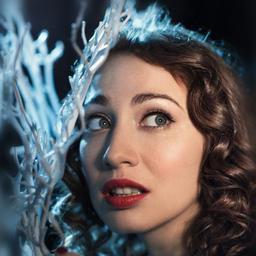}
				\includegraphics[width=1.13in]{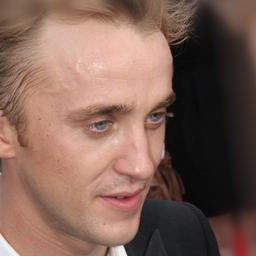}
				\includegraphics[width=1.13in]{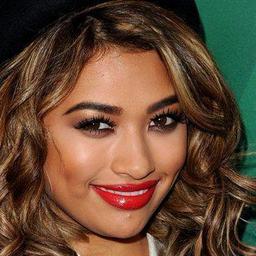}
				\includegraphics[width=1.13in]{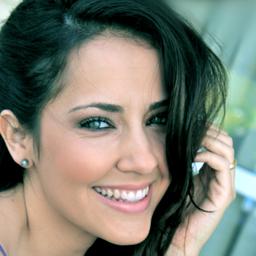}
				\includegraphics[width=1.13in]{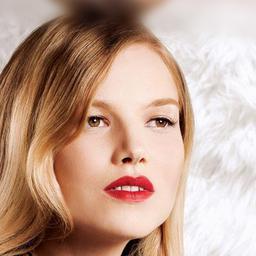}
				\includegraphics[width=1.13in]{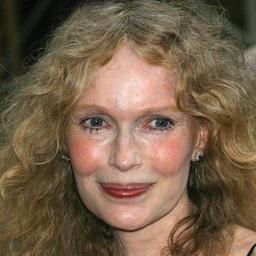}
				\includegraphics[width=1.13in]{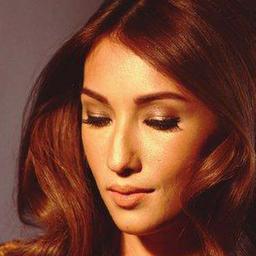}
			\end{minipage}
		}
		\subfigure[Masked GT]{
			\begin{minipage}[t]{0.15\linewidth}
				\includegraphics[width=1.13in]{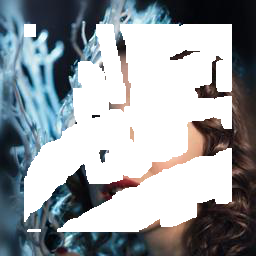}
				\includegraphics[width=1.13in]{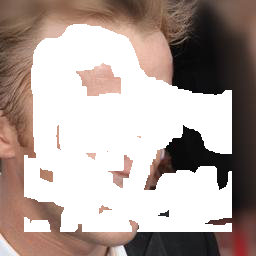}
				\includegraphics[width=1.13in]{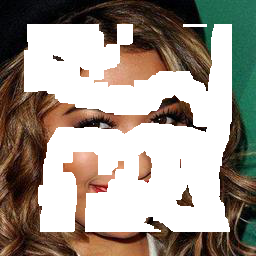}
				\includegraphics[width=1.13in]{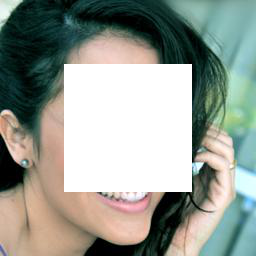}
				\includegraphics[width=1.13in]{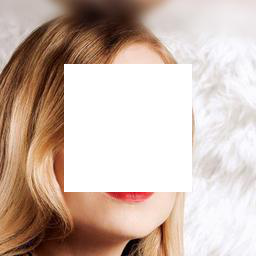}
				\includegraphics[width=1.13in]{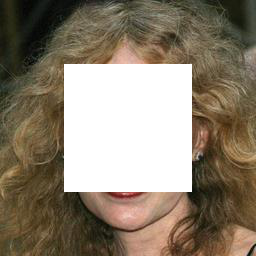}
				\includegraphics[width=1.13in]{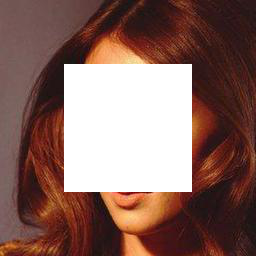}
			\end{minipage}
		}
		\subfigure[CA]{
			\begin{minipage}[t]{0.15\linewidth}
				\includegraphics[width=1.13in]{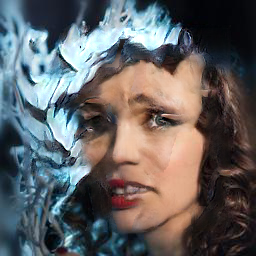}
				\includegraphics[width=1.13in]{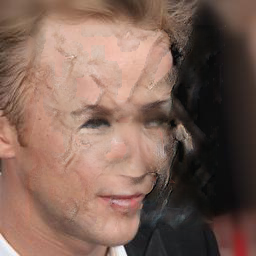}
				\includegraphics[width=1.13in]{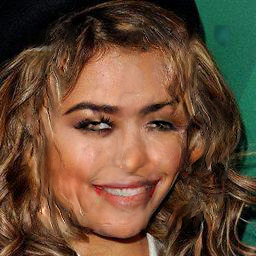}
				\includegraphics[width=1.13in]{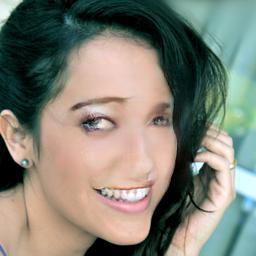}
				\includegraphics[width=1.13in]{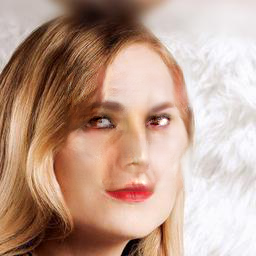}
				\includegraphics[width=1.13in]{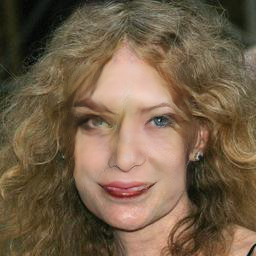}
				\includegraphics[width=1.13in]{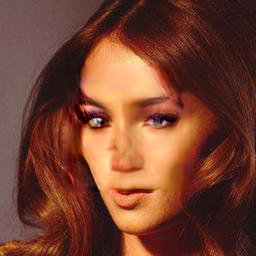}
			\end{minipage}
		}
		\subfigure[EC]{
			\begin{minipage}[t]{0.15\linewidth}
				\includegraphics[width=1.13in]{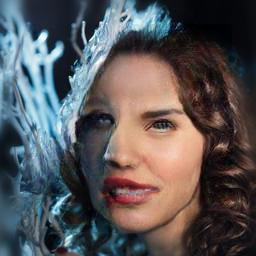}
				\includegraphics[width=1.13in]{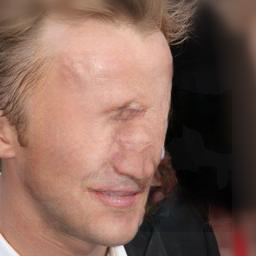}
				\includegraphics[width=1.13in]{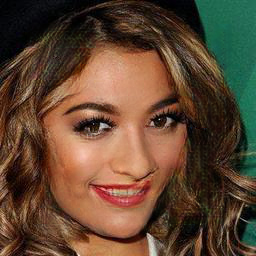}
				\includegraphics[width=1.13in]{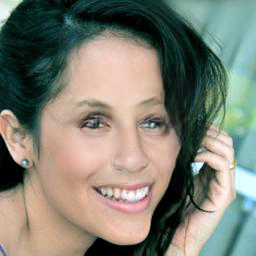}
				\includegraphics[width=1.13in]{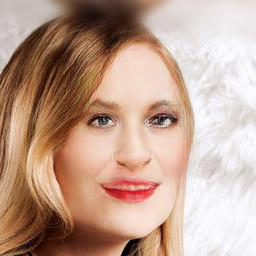}
				\includegraphics[width=1.13in]{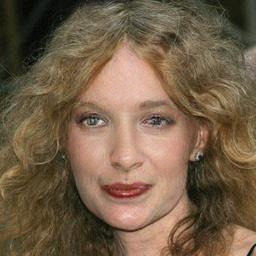}
				\includegraphics[width=1.13in]{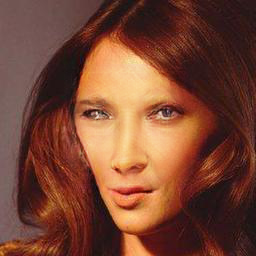}

			\end{minipage}
		}
		\subfigure[PIC]{
			\begin{minipage}[t]{0.15\linewidth}
				\includegraphics[width=1.13in]{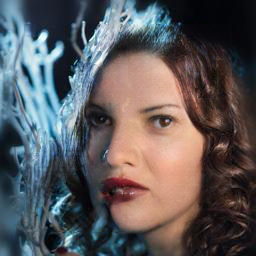}
				\includegraphics[width=1.13in]{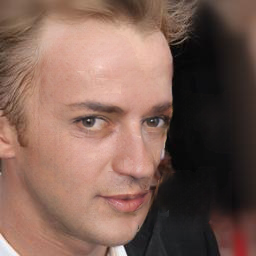}
				\includegraphics[width=1.13in]{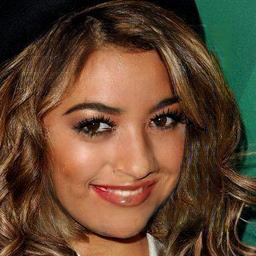}
				\includegraphics[width=1.13in]{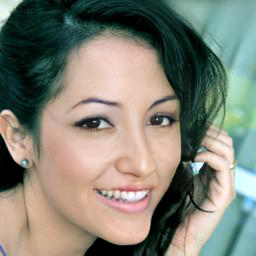}
				\includegraphics[width=1.13in]{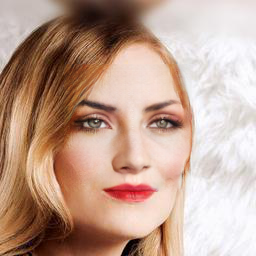}
				\includegraphics[width=1.13in]{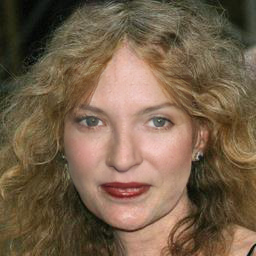}
				\includegraphics[width=1.13in]{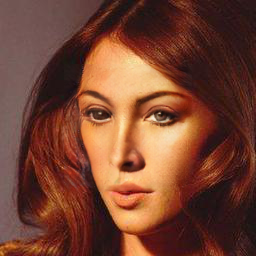}
			\end{minipage}
		}
		\subfigure[Ours]{
			\begin{minipage}[t]{0.15\linewidth}
				\includegraphics[width=1.13in]{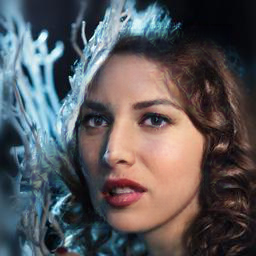}
				\includegraphics[width=1.13in]{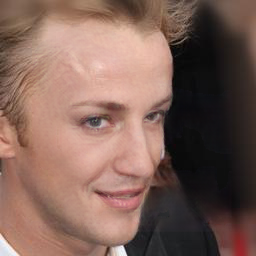}
				\includegraphics[width=1.13in]{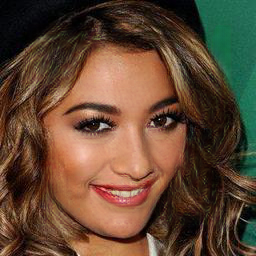}
				\includegraphics[width=1.13in]{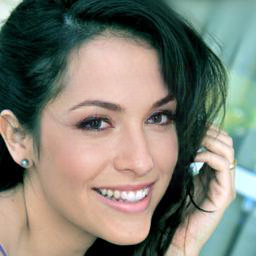}
				\includegraphics[width=1.13in]{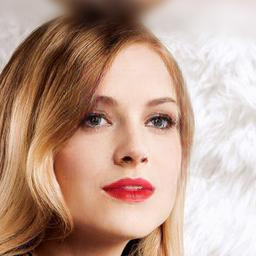}
				\includegraphics[width=1.13in]{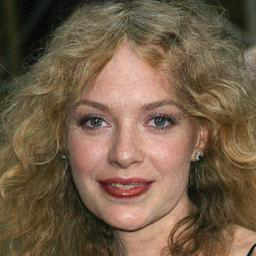}
				\includegraphics[width=1.13in]{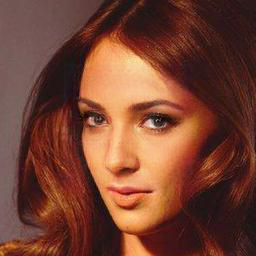}
			\end{minipage}
		}
	\end{center}
\vspace{-7pt}
	\caption{More results with other state-of-the-art techniques on the CelebA-HQ dataset. (a) shows the ground-truth images. (b) depicts the masked versions of (a). (c)-(f) are the results obtained by CA, EC, PIC, and our LaFIn, respectively. }
		\vspace{-7pt}
		\label{fig:more2}
\end{figure*}

\begin{figure*}[ht]
	\begin{center}
		\subfigure[Ground truth (GT)]{
			\begin{minipage}[t]{0.15\linewidth}
				\includegraphics[width=1.13in]{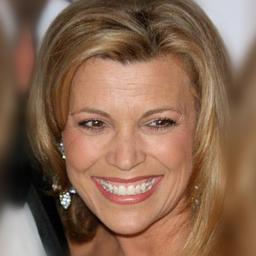}
				\includegraphics[width=1.13in]{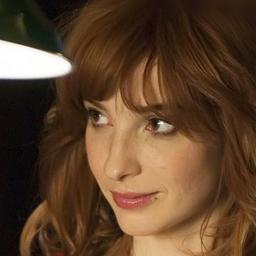}
				\includegraphics[width=1.13in]{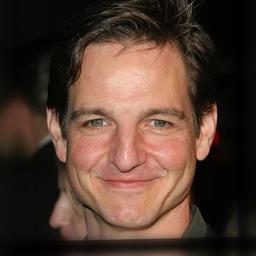}
				\includegraphics[width=1.13in]{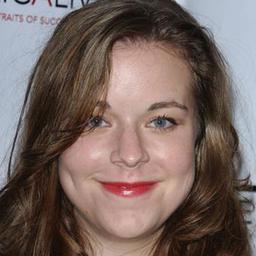}
				\includegraphics[width=1.13in]{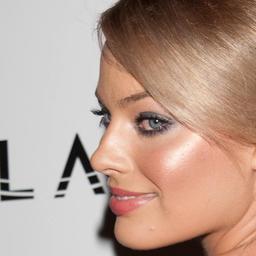}
				\includegraphics[width=1.13in]{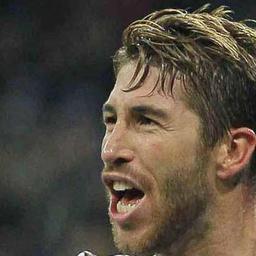}
				\includegraphics[width=1.13in]{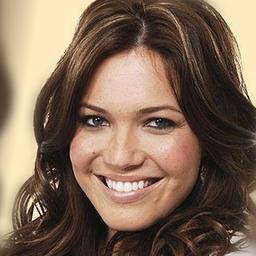}
			\end{minipage}
		}
		\subfigure[Masked GT]{
			\begin{minipage}[t]{0.15\linewidth}
				\includegraphics[width=1.13in]{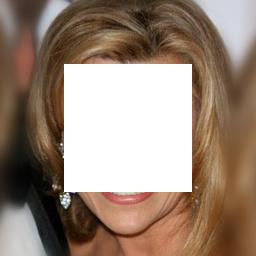}
				\includegraphics[width=1.13in]{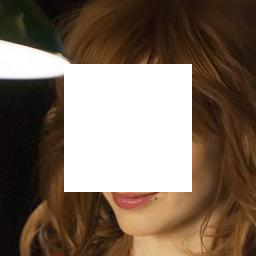}
				\includegraphics[width=1.13in]{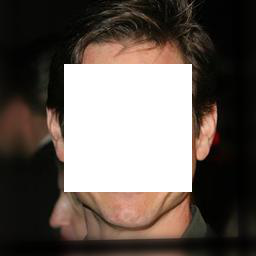}
				\includegraphics[width=1.13in]{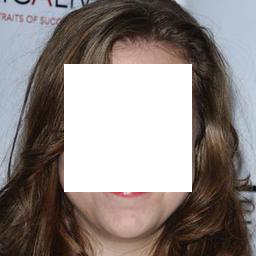}
				\includegraphics[width=1.13in]{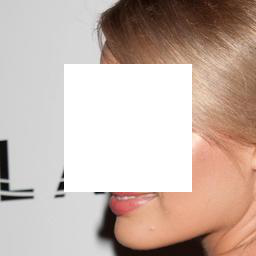}
				\includegraphics[width=1.13in]{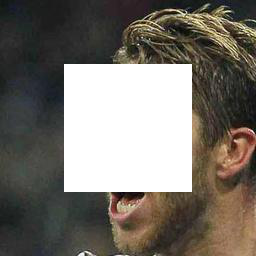}
				\includegraphics[width=1.13in]{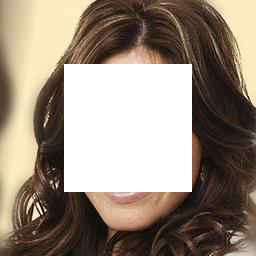}
			\end{minipage}
		}
		\subfigure[CA]{
			\begin{minipage}[t]{0.15\linewidth}
				\includegraphics[width=1.13in]{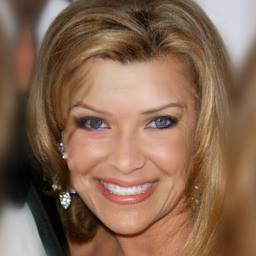}
				\includegraphics[width=1.13in]{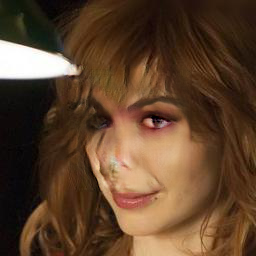}
				\includegraphics[width=1.13in]{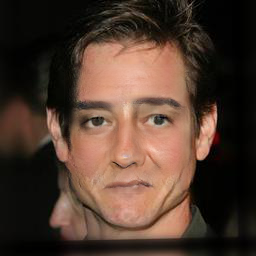}
				\includegraphics[width=1.13in]{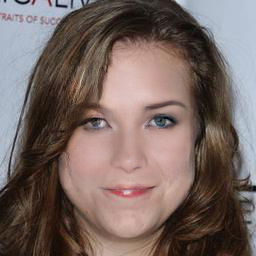}
				\includegraphics[width=1.13in]{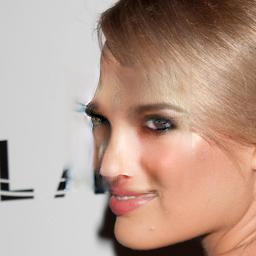}
				\includegraphics[width=1.13in]{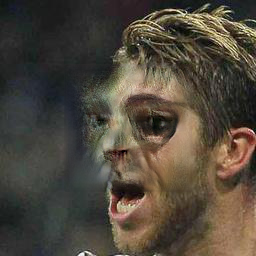}
				\includegraphics[width=1.13in]{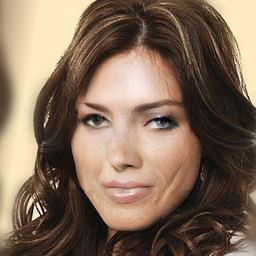}
			\end{minipage}
		}
		\subfigure[EC]{
			\begin{minipage}[t]{0.15\linewidth}
				\includegraphics[width=1.13in]{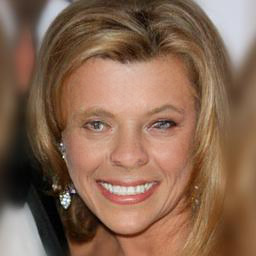}
				\includegraphics[width=1.13in]{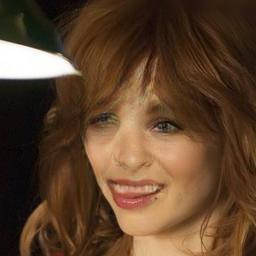}
				\includegraphics[width=1.13in]{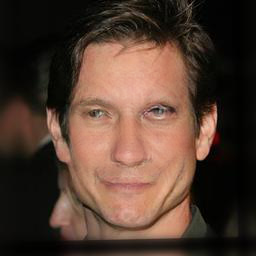}
				\includegraphics[width=1.13in]{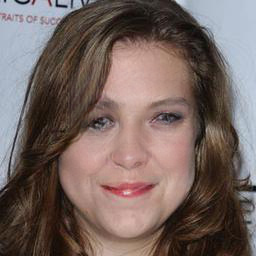}
				\includegraphics[width=1.13in]{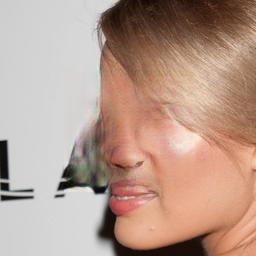}
				\includegraphics[width=1.13in]{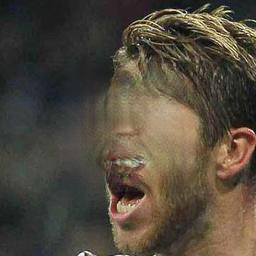}
				\includegraphics[width=1.13in]{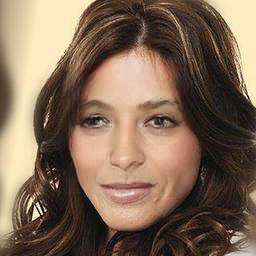}

			\end{minipage}
		}
		\subfigure[PIC]{
			\begin{minipage}[t]{0.15\linewidth}
				\includegraphics[width=1.13in]{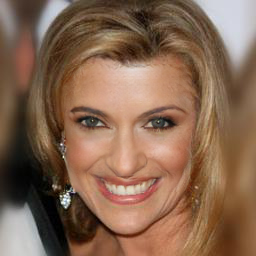}
				\includegraphics[width=1.13in]{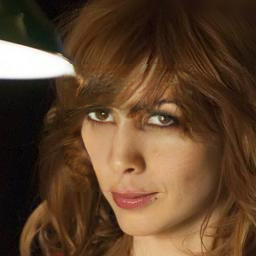}
				\includegraphics[width=1.13in]{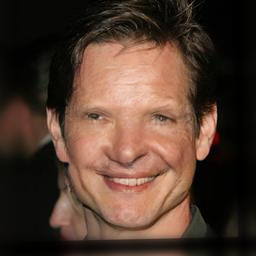}
				\includegraphics[width=1.13in]{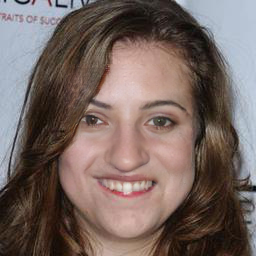}
				\includegraphics[width=1.13in]{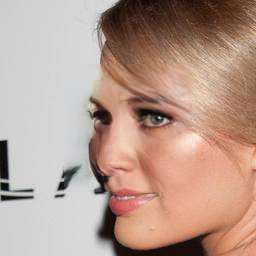}
				\includegraphics[width=1.13in]{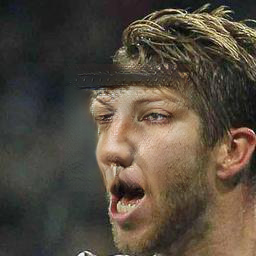}
				\includegraphics[width=1.13in]{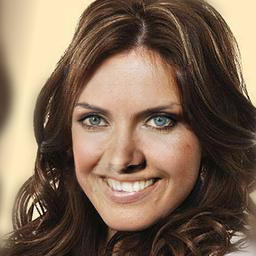}
			\end{minipage}
		}
		\subfigure[Ours]{
			\begin{minipage}[t]{0.15\linewidth}
				\includegraphics[width=1.13in]{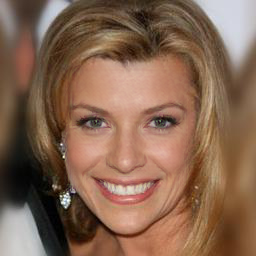}
				\includegraphics[width=1.13in]{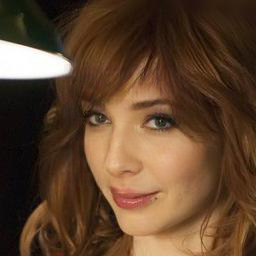}
				\includegraphics[width=1.13in]{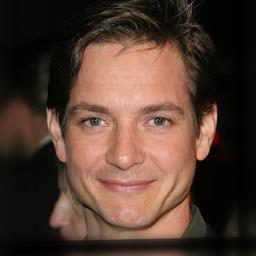}
				\includegraphics[width=1.13in]{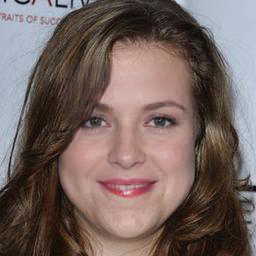}
				\includegraphics[width=1.13in]{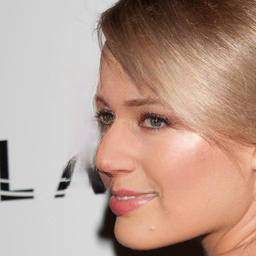}
				\includegraphics[width=1.13in]{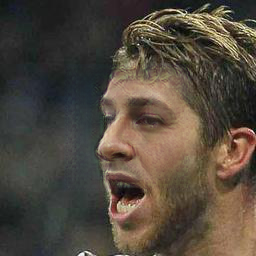}
				\includegraphics[width=1.13in]{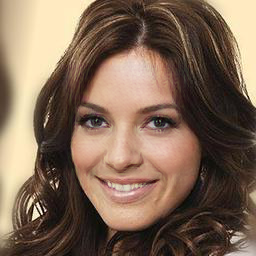}
			\end{minipage}
		}
	\end{center}
\vspace{-7pt}
	\caption{More results with other state-of-the-art techniques on the CelebA-HQ dataset. (a) shows the ground-truth images. (b) depicts the masked versions of (a). (c)-(f) are the results obtained by CA, EC, PIC, and our LaFIn, respectively. }
		\vspace{-7pt}
		\label{fig:more3}
\end{figure*}

\begin{figure*}[ht]
	\begin{center}
		\subfigure[Ground truth (GT)]{
			\begin{minipage}[t]{0.15\linewidth}
				\includegraphics[width=1.13in]{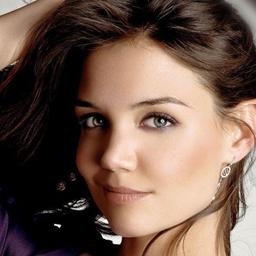}
				\includegraphics[width=1.13in]{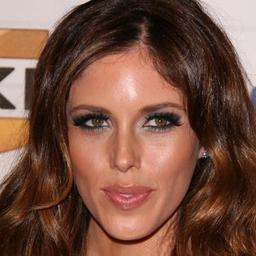}
				\includegraphics[width=1.13in]{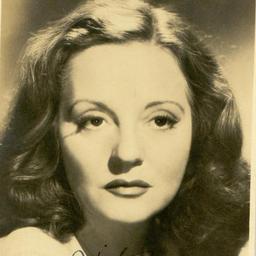}
				\includegraphics[width=1.13in]{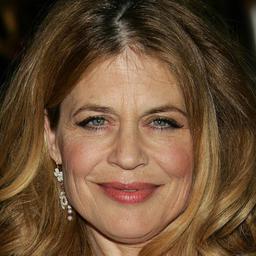}
				\includegraphics[width=1.13in]{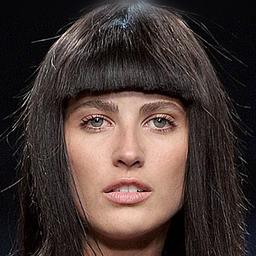}
				\includegraphics[width=1.13in]{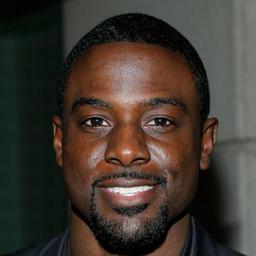}
				\includegraphics[width=1.13in]{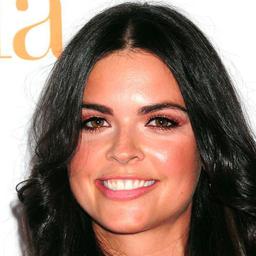}
			\end{minipage}
		}
		\subfigure[Masked GT]{
			\begin{minipage}[t]{0.15\linewidth}
				\includegraphics[width=1.13in]{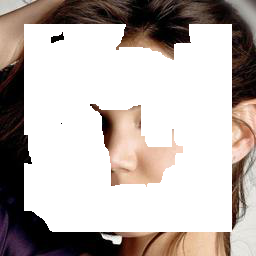}
				\includegraphics[width=1.13in]{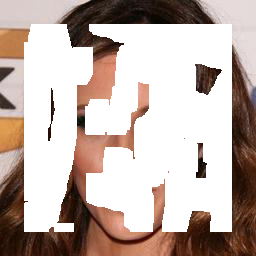}
				\includegraphics[width=1.13in]{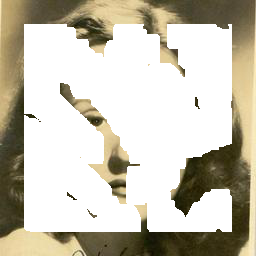}
				\includegraphics[width=1.13in]{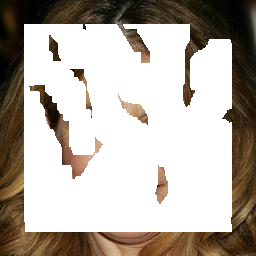}
				\includegraphics[width=1.13in]{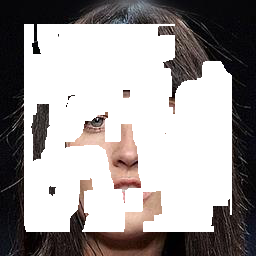}
				\includegraphics[width=1.13in]{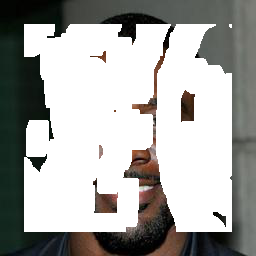}
				\includegraphics[width=1.13in]{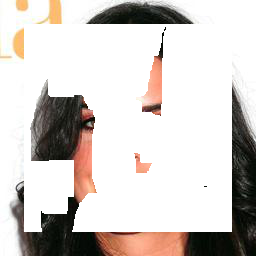}
			\end{minipage}
		}
		\subfigure[CA]{
			\begin{minipage}[t]{0.15\linewidth}
				\includegraphics[width=1.13in]{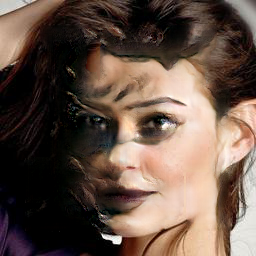}
				\includegraphics[width=1.13in]{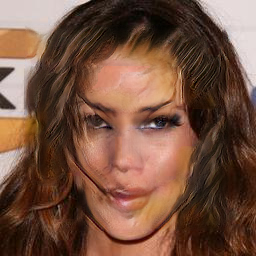}
				\includegraphics[width=1.13in]{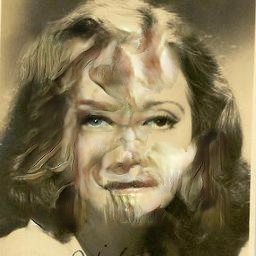}
				\includegraphics[width=1.13in]{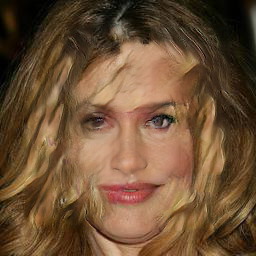}
				\includegraphics[width=1.13in]{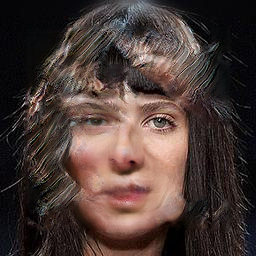}
				\includegraphics[width=1.13in]{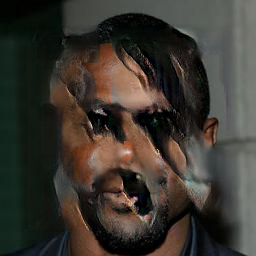}
				\includegraphics[width=1.13in]{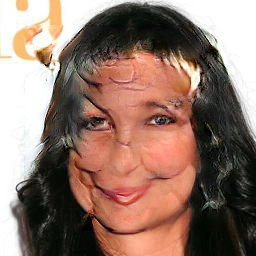}	
			\end{minipage}
		}
		\subfigure[EC]{
			\begin{minipage}[t]{0.15\linewidth}
				\includegraphics[width=1.13in]{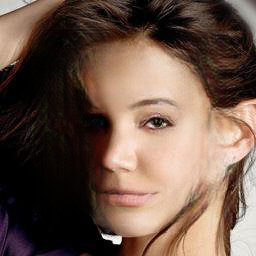}
				\includegraphics[width=1.13in]{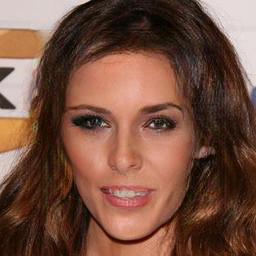}
				\includegraphics[width=1.13in]{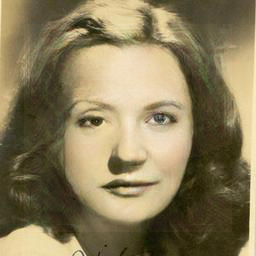}
				\includegraphics[width=1.13in]{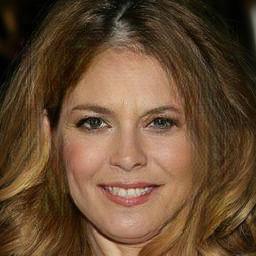}
				\includegraphics[width=1.13in]{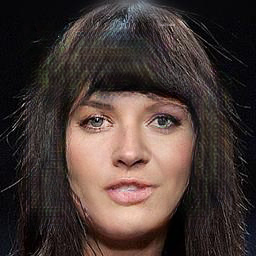}
				\includegraphics[width=1.13in]{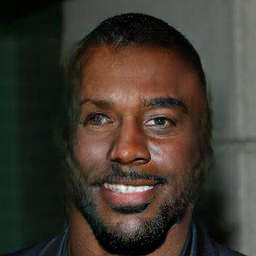}
				\includegraphics[width=1.13in]{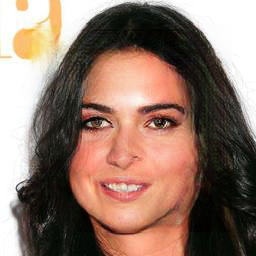}
			\end{minipage}
		}
		\subfigure[PIC]{
			\begin{minipage}[t]{0.15\linewidth}
				\includegraphics[width=1.13in]{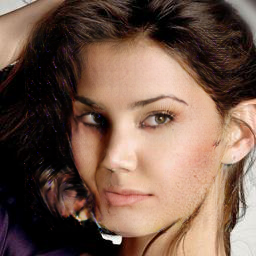}
				\includegraphics[width=1.13in]{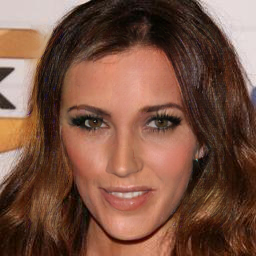}
				\includegraphics[width=1.13in]{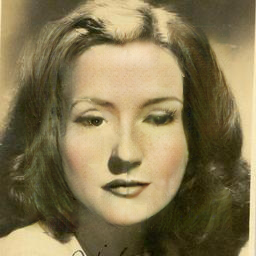}
				\includegraphics[width=1.13in]{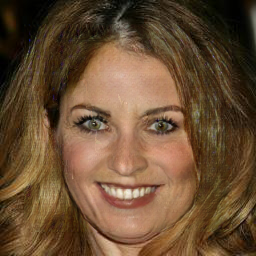}
				\includegraphics[width=1.13in]{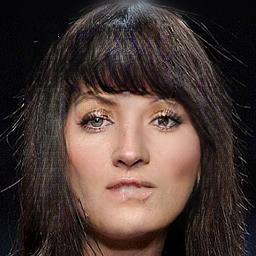}
				\includegraphics[width=1.13in]{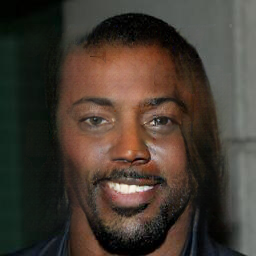}
				\includegraphics[width=1.13in]{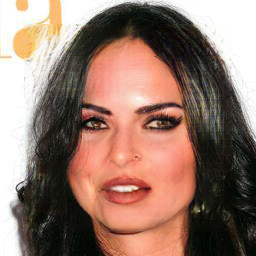}

			\end{minipage}
		}
		\subfigure[Ours]{
			\begin{minipage}[t]{0.15\linewidth}
				\includegraphics[width=1.13in]{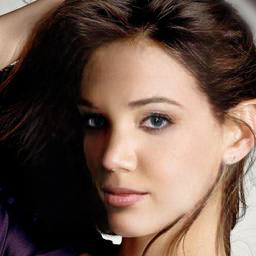}
				\includegraphics[width=1.13in]{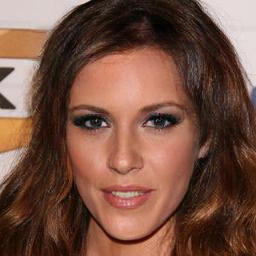}
				\includegraphics[width=1.13in]{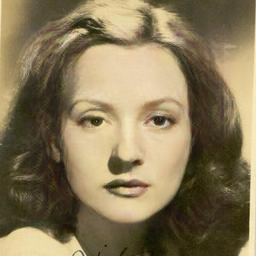}
				\includegraphics[width=1.13in]{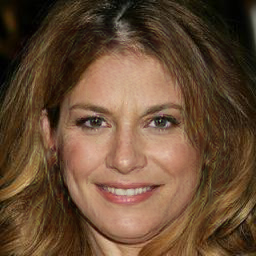}
				\includegraphics[width=1.13in]{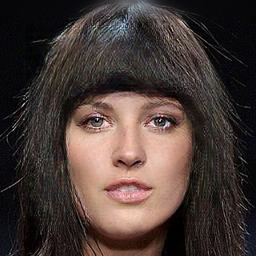}
				\includegraphics[width=1.13in]{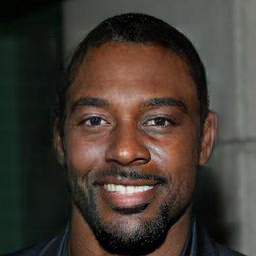}
				\includegraphics[width=1.13in]{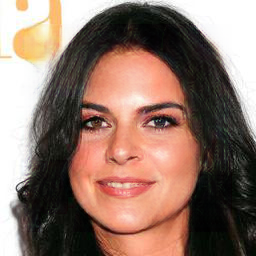}
				
			\end{minipage}
		}
	\end{center}
\vspace{-7pt}
	\caption{More results with other state-of-the-art techniques on the CelebA-HQ dataset. (a) shows the ground-truth images. (b) depicts the masked versions of (a). (c)-(f) are the results obtained by CA, EC, PIC, and our LaFIn, respectively. }
		\vspace{-7pt}
		\label{fig:more4}
\end{figure*}

\begin{figure*}[h]
		\begin{center}
		\subfigure[Ground truth (GT)]{
			\begin{minipage}[t]{0.15\linewidth}								
				\includegraphics[width=1.13in]{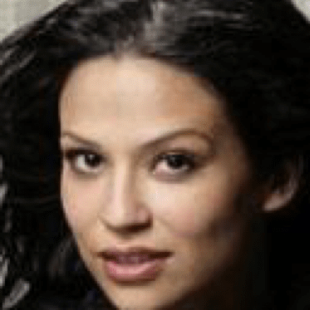}
				\includegraphics[width=1.13in]{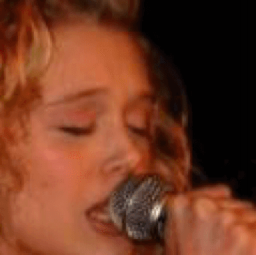}
			\end{minipage}
		}
		\subfigure[Masked GT]{
			\begin{minipage}[t]{0.15\linewidth}
				\includegraphics[width=1.13in]{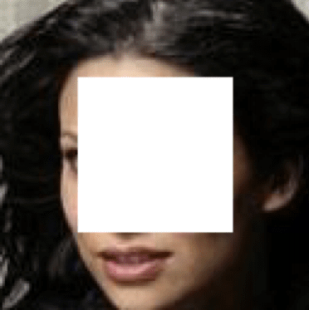}
				\includegraphics[width=1.13in]{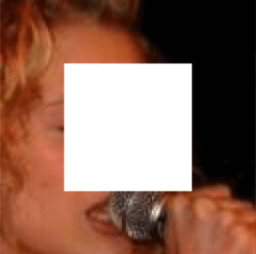}
			\end{minipage}
		}
		\subfigure[CE]{
			\begin{minipage}[t]{0.15\linewidth}
				\includegraphics[width=1.13in]{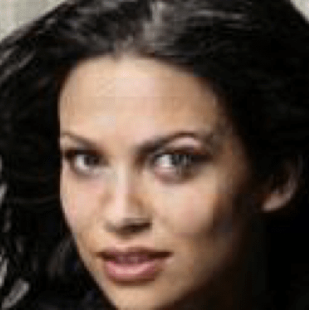}
				\includegraphics[width=1.13in]{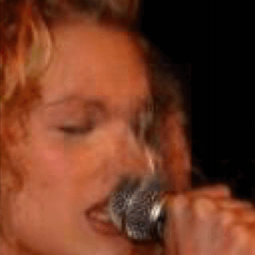}
			\end{minipage}
		}
		\subfigure[GFC]{
			\begin{minipage}[t]{0.15\linewidth}
				\includegraphics[width=1.13in]{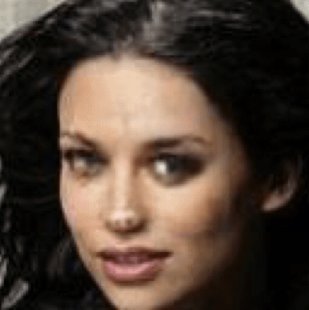}
				\includegraphics[width=1.13in]{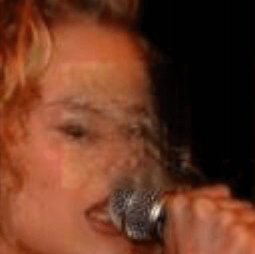}
			\end{minipage}
		}
		\subfigure[GAFC]{
			\begin{minipage}[t]{0.15\linewidth}
				\includegraphics[width=1.13in]{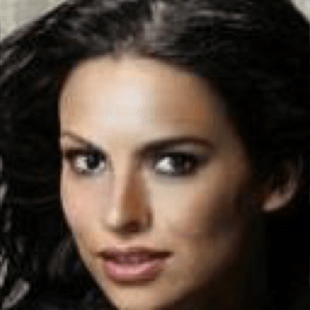}
				\includegraphics[width=1.13in]{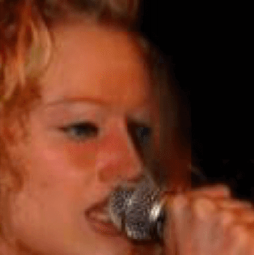}
			\end{minipage}
		}
		\subfigure[Ours]{
			\begin{minipage}[t]{0.15\linewidth}
				\includegraphics[width=1.13in]{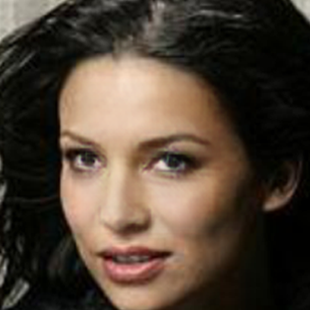}
				\includegraphics[width=1.13in]{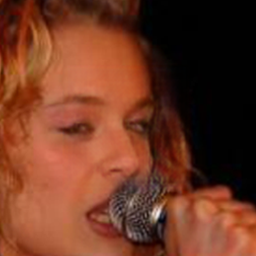}
			\end{minipage}
		}
		
	\end{center}
	\vspace{-5pt}
	\caption{Visual comparison between the competitors on the CelebA dataset. (a) shows the ground-truth images. (b) depicts the masked versions of (a). (c)-(f) are the results obtained by CE, GFC, GAFC, and our LaFIn, respectively.}
		\vspace{-7pt}
	\label{fig:more5}
\end{figure*}

In Table 1 of the main paper, our LaFIn falls behind PIC in terms of FID $6.63$ \textit{vs.} $4.98$ in the case of center mask. First we give the definition of FID as follows:
\begin{equation}
{\rm FID}(x,g) = \|\mu_x-\mu_g\|_2^2 + {\rm Tr}(\Sigma_x+\Sigma_g-2(\Sigma_x\Sigma_g)^{\frac{1}{2}}).
\label{eq1}
\end{equation}
Assuming that the extracted features $x$ and $g$ follow multidimensional Gaussian distributions $({\bf \mu}_x,{\bf \Sigma}_x)$ and $({\bf \mu}_g, {\bf \Sigma}_g)$ respectively, the FID calculates the Frechet distance between the two distributions. And ${\rm Tr}$ stands for the trace of matrix. From Eq.(\ref{eq1}) we can see that the FID takes both the mean and the variance of features into consideration. As Figures \ref{fig:more2} and \ref{fig:more3} show, in the situation of center mask, the available information in images for inpainting is limited and our LaFIn tends to generate common but reasonable results, which decreases the performance in terms of FID, especially in the variance term. While PIC is designed to generate pluralistic features, but some of the results are not visually satisfactory. More results comparing with CA \cite{yu2018generative}, EC\cite {nazeri2019edgeconnect}, PIC\cite {zheng2019pluralistic} on CelebA-HQ and CE\cite{CE}, GFC \cite{GFC}, GAFC \cite{GAFC} on CelebA are shown in Figure \ref{fig:more1} to Figure \ref{fig:more5}.

\end{appendices}

{
\bibliographystyle{ieee}
\bibliography{references}
}

\end{document}